\documentclass{bmvc2k}
\usepackage[group-separator={,}]{siunitx}
\usepackage{booktabs}

\title{Play and Learn: Using Video Games to Train Computer Vision Models}

\addauthor{Alireza Shafaei}{http://cs.ubc.ca/~shafaei}{1}
\addauthor{James J. Little}{http://cs.ubc.ca/~little}{1}
\addauthor{Mark Schmidt}{http://cs.ubc.ca/~schmidtm}{1}

\addinstitution{
 Department of Computer Science\\
 University of British Columbia\\
 Vancouver, Canada
}

\runninghead{Shafaei \bmvaEtAl}{Using Video Games to Train Computer Vision Models}

\def\ie{\emph{i.e}\bmvaOneDot}
\def\etal{\emph{et al}\bmvaOneDot}

\graphicspath{{./images/}}
\begin{document}
\newgeometry{twoside,headsep=3mm,papersize={410pt,620pt},inner=14mm,outer=6mm,top=3mm,includehead,bottom=5mm,heightrounded}

\maketitle

\begin{abstract}
Video games are a compelling source of annotated data as they can readily provide fine-grained groundtruth for diverse tasks.
However, it is not clear whether the synthetically generated data has enough resemblance to the real-world images to improve the performance of computer vision models in practice.
We present experiments assessing the effectiveness on real-world data of systems trained on synthetic RGB images that are extracted from a video game.
We collected over \num{60000} synthetic samples from a modern video game with similar conditions to the real-world CamVid and Cityscapes datasets.
We provide several experiments to demonstrate that the synthetically generated RGB images can be used to improve the performance of deep neural networks on both image segmentation and depth estimation.
These results show that a convolutional network trained on synthetic data achieves a similar test error to a network that is trained on real-world data for dense image classification. Furthermore, the synthetically generated RGB images can provide similar or better results compared to the real-world datasets if a simple domain adaptation technique is applied.
Our results suggest that collaboration with game developers for an accessible interface to gather data is potentially a fruitful direction for future work in computer vision.
\end{abstract}

\section{Introduction}
\label{sec:intro}

Deep neural networks have been setting new records in almost all of the computer vision challenges and continue to grow in the broader field of artificial intelligence.
One of the key components of a successful deep network recipe is the availability of a sufficiently large dataset for training and evaluation.
The Imagenet Large Scale Visual Recognition Challenge (ILSVRC)~\cite{ILSVRC15}, with over 1 million training images, has been one of the primary benchmarks for evaluation of several deep network architectures over the past few years. As architectures such as Resnet~\cite{he2015deep} approach a 3.57\% top-5 error on the Imagenet classification task, it is likely that a larger and more complex dataset will become necessary soon.
However, compiling datasets has its difficulties -- depending on the desired granularity of the groundtruth, the costs and the labor could grow rather quickly.
\newgeometry{twoside,headsep=3mm,papersize={410pt,620pt},inner=9mm,outer=6mm,top=3mm,includehead,bottom=5mm,heightrounded}

Meanwhile, to cope with limited available data, there are several approaches to the training of deep networks. The common practices include data augmentation through various image transformations~\cite{krizhevsky2012imagenet} or use of pre-trained networks that are trained for a similar task as a warm-start in the original optimization problem.
A more radical approach is to synthesize training samples through computer-generated (CG) imagery.

One of the main concerns in the use of CG data for the training is whether the source domain is close enough to the target domain (\ie, the domain of real-world images) to make an effective application possible. For depth images of the Kinect sensors, for instance, several lines of work~\cite{shotton2013efficient,tompson2014real,rogez2015first,Shafaei2016} demonstrate the efficacy and practical similarities between computer-generated depth images and the output of consumer level depth sensors. There are also methods that rely on indirect representations such as the HOG templates~\cite{hejrati2014analysis} or geometrical descriptors~\cite{lim2014fpm,stark2010back}, that can be efficiently synthesized and used. For the direct synthesis and application of RGB data, however, there is a limited amount of work that only investigates simple rendering of 3d CAD models~\cite{Peng_2015_ICCV,aubry2015understanding,lim2013parsing,Aubry_2014_CVPR,sun2014virtual} or synthetic text~\cite{gupta2016synthetic}.
The question of interest is whether existing video games with photorealistic environments can be a useful source of synthetic RGB data to address computer vision tasks.

Although video games generate images from a finite set of textures, there is variation in viewpoint, illumination, weather, and level of detail which could provide valuable augmentation of the data. In addition to full control over the environment, video games can also provide us with groundtruth data such as dense image class annotations, depth information, radiance, irradiance, and reflectance which may not be straightforward, or even possible, to collect from real data. Other measures, such as the precise location of the game character within the environment, could be useful for development of visual SLAM algorithms.

In this work, we focus our attention on the RGB domain of a modern video game and run various experiments to gauge the efficacy of using CG RGB data directly for computer vision problems. We collect over \num{60000} outdoor images under conditions similar to the CamVid~\cite{Brostow2009} and the Cityscapes~\cite{Cordts2016Cityscapes} datasets and present experiments on two computer vision problems: (i)~dense image annotation, and (ii)~depth estimation from RGB. We show that a convolutional network trained on synthetic data achieves a similar test error to a network that is trained on real-world data. Furthermore, after fine-tuning, our results show a network that is pre-trained on synthetic data can outperform a network that is pre-trained on real-world data.
\section{Previous Work}
\label{sec:prev_work}

\textbf{Synthetic Data.}
Synthetic data has a successful history in computer vision. Taylor~\etal{}~\cite{taylor2007ovvv} present a system called ObjectVideo Virtual Video (OVVV) based on Half-life~\cite{hl2} for evaluation of tracking in surveillance systems. Marin~\etal{}~\cite{marin2010learning} 
extend OVVV to perform pedestrian detection with HOG~\cite{dalal2005histograms} features. In the recent literature, a variety of methods~\cite{lim2013parsing,Peng_2015_ICCV,sun2014virtual,Aubry_2014_CVPR} use 3d CAD models with simple rendering to tackle vision problems. Peng~\etal~\cite{Peng_2015_ICCV}, and Sun and Saenko~\cite{sun2014virtual} use non-photorealistic 3d CAD models to improve object detection. Lim~\etal~\cite{lim2013parsing}, and Aubry~\etal~\cite{Aubry_2014_CVPR} use CAD models for detection and object alignment in the image.
Aubry and Russell~\cite{aubry2015understanding} use synthetic RGB images rendered from CAD models to analyze the response pattern and the behavior of neurons in the commonly used deep convolutional networks. 
Rematas~\etal~\cite{rematas2014image} use 3d models to synthesize novel viewpoints of objects in real world images. Stark~\etal~\cite{stark2010back}, Lim~\etal~\cite{lim2014fpm}, and Liebelt and Schmid~\cite{liebelt2010multi} learn intermediate geometric descriptors from 3d models to perform object detection. 
Butler \etal~\cite{Butler:ECCV:2012} present the synthetic Sintel dataset for evaluation of optical flow methods. 
Synthetic depth images are also successfully used for human pose estimation~\cite{shotton2013efficient,Shafaei2016} and hand pose estimation~\cite{tompson2014real,rogez2015first}. Kaneva~\etal~\cite{kaneva2011evaluation} study robustness of image features under viewpoint and illumination changes in a photorealistic virtual world.
In contrast to previous work, we take a different approach to synthetic models. Instead of rendering simple 3d CAD models in isolation, we take a step further and collect synthetic data in a simulated photorealistic world within the broad context of street scenes. We are specifically targeting the use of modern video games to generate densely annotated groundtruth to train computer vision models.

\textbf{Dense Image Classification}.
Deep Convolutional Networks are extensively used for dense image segmentation~\cite{YuKoltun2016,long2015fully,Papandreou2015,crfasrnn_iccv2015}.
The fully convolutional networks of Long~\etal~\cite{long2015fully} is among the first to popularize deep network architectures that densely label input images.
Zheng \etal~\cite{crfasrnn_iccv2015} build on top of the architecture in Long \etal~\cite{long2015fully} and integrate CRFs with Gaussian pairwise potentials to yield a significant gain in image segmentation. The current state-of-the-art methods, such as the one presented by Liu \etal~\cite{Liu_2015_ICCV}, use variants of fully convolutional networks as a building block on top of which different recurrent neural networks or graphical models are proposed. We use the basic fully convolutional architecture of Long~\etal~\cite{long2015fully} as it provides the basis of the follow-up developments in this area.

\textbf{Depth Estimation from RGB.} One of the early studies on unconstrained depth estimation from single RGB images is the work of Saxena~\etal~\cite{saxena20083} in which the authors present a hierarchical Markov random field to estimate the depth.
More recently, Zhuo \etal~\cite{Zhuo_2015_CVPR} present an energy minimization problem that incorporates semantic information at multiple levels of abstraction to generate a depth estimate.
Li \etal~\cite{li2015depth} and Liu \etal~\cite{Liu_2015_CVPR} use deep networks equipped with a conditional random field that estimates the depth.
More recently, Eigen and Fergus~\cite{EigenICCV15} presented a multi-scale deep convolutional architecture that can predict depth, normals, and dense labels.
Instead of regressing against the metric data directly, Zoran \etal~\cite{Zoran_2015_ICCV} propose a general method to estimate reflectance, shading, and depth by learning a model to predict ordinal relationships. The input image is first segmented into SLIC superpixels~\cite{achanta2012slic} on top of which a multi-scale neighbourhood graph is constructed. Zoran \etal~\cite{Zoran_2015_ICCV} use the neighbourhood graph and generate ordinal queries on its edges, and then use the results to construct a globally consistent ranking by solving a quadratic program. We apply the method of Zoran~\etal~\cite{Zoran_2015_ICCV} in our study and show improvement in the depth estimation task through the use of synthetic data.

\textbf{Transfer Learning.} A closely related area of work to our study is transfer learning (see Pan and Yang~\cite{pan2010survey} for a review). The early studies of transfer learning with Deep Convolutional Networks successfully demonstrated domain adaptation through pre-training on source data and fine-tuning on target data~\cite{yosinski_2014_NIPS,sharif2014cnn,donahue2014decaf}. Further studies such as the work of Ganin~\etal~\cite{ganin_jmlr} present more sophisticated approaches to domain adaptation. In this work, we apply the most widely used fine-tuning approach to domain adaptation and leave further studies on feature transferability of the synthetic data to the future work.

\section{Datasets}
\label{sec:Datasets}
In this work, we focus on outdoor on-the-road datasets as we can easily collect this data from the video game.
For real-world images, there are a few available datasets~\cite{Brostow2009,Cordts2016Cityscapes,scharwachter2013efficient,geiger2013vision} of which we use Cityscapes~\cite{Cordts2016Cityscapes} and CamVid~\cite{Brostow2009}. The selected datasets provide the most appropriate setting regarding the available groundtruth and visual similarity.
\subsection{VG Dataset}
\label{sec:VG_dataset}
The dataset consists of over \num{60000} frames collected from a video game\footnote{Due to legal concerns we were asked to not disclose the title of the game.}.
To gather this data we use a camera on the hood of a car, similar to the configuration of the CamVid~\cite{Brostow2009} or Cityscapes~\cite{Cordts2016Cityscapes} datasets.
The weather is kept fixed at sunny and the time of the day is fixed at 11:00 AM.
This atmospheric setting was chosen to make the synthetic data similar to the real-world data, although note that a key advantage of \texttt{VG} is that it would be easy to sample data under non-ideal conditions (while it might be impossible to collect reliable real data under many conditions).
The image resolution of the game is $1024\times768$ with the highest possible graphics configuration.
The autonomous driver randomly drives around the city while obeying the traffic laws. Every second, a sample data is collected from the game.
Each sample contains the RGB image, groundtruth semantic segmentation, depth image, and the surface normals.
The groundtruth semantic segmentation that we were able to automatically extract from the game is over the label set $\{\textrm{Sky},\, \textrm{Pedestrian},\, \textrm{Cars},\, \textrm{Trees}\}$. The help of the game developers is likely to be required in order to extract more labels.

\begin{figure}
    \centering
    \includegraphics[width=0.4\textwidth]{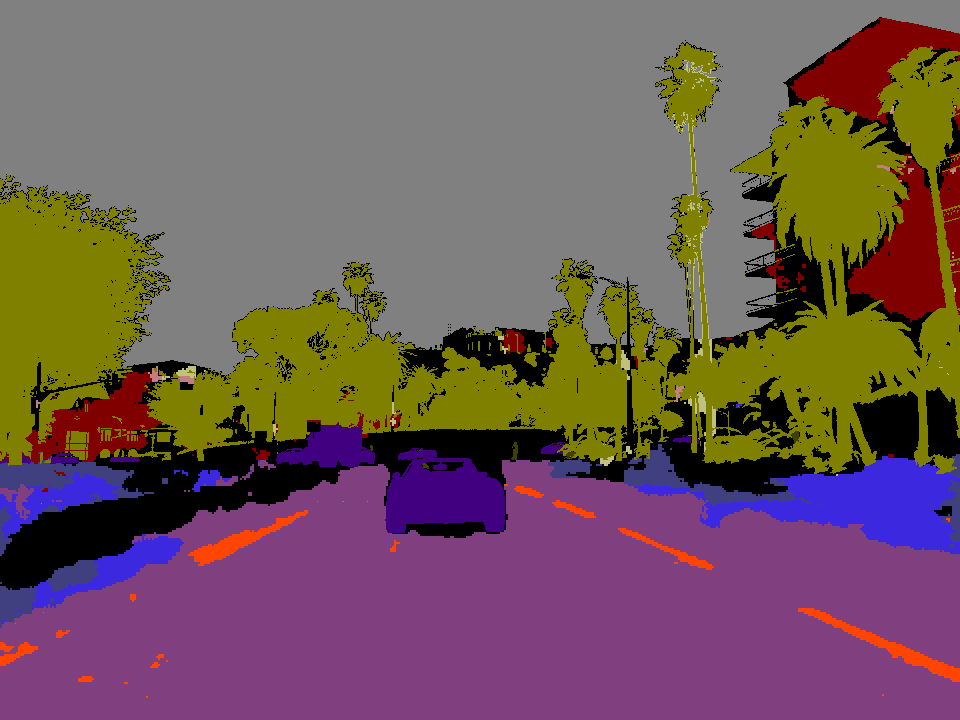}
    \includegraphics[width=0.4\textwidth]{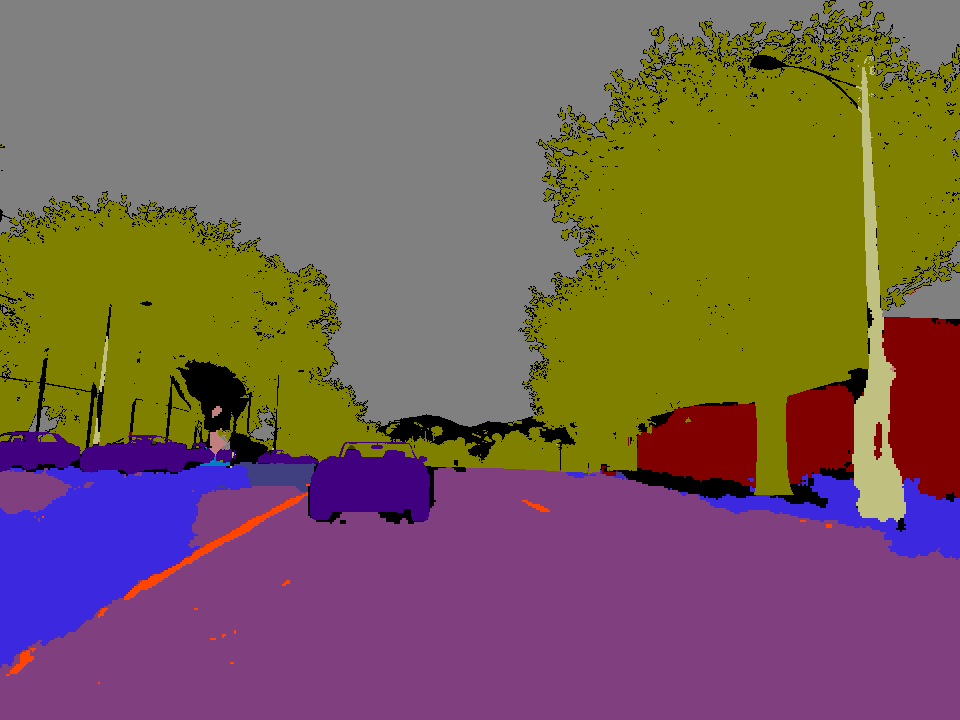}
    \caption{Densely labeled samples from the \texttt{VG+} dataset. The label space of this dataset is the same as the CamVid~\cite{Brostow2009} dataset.}
    \label{fig:sample_dataset2}
\end{figure}

We also consider a label-augmented version of the \texttt{VG} dataset, which we call \texttt{VG+}. To augment the label space with additional classes, we use SegNet~\cite{kendall2015bayesian}, one of the top performing methods on CamVid~\cite{Brostow2009} dataset with available code and data, to classify the images of the \texttt{VG}. We then refine the labels with the true labels of the \texttt{VG} dataset and clean-up the data automatically using the depth and the surface normals.
See Fig.~\ref{fig:sample_dataset2} for samples.

Note that, unlike the groundtruth dense annotations of \texttt{VG}, the groundtruth label space $\mathcal{Y}$ in \texttt{VG+} is noisy and may partially exhibit the biases of the SegNet~\cite{kendall2015bayesian}. However, the input space $\mathcal{X}$ remains the same. Consequently, the results that are derived from \texttt{VG+} are weaker but still can provide us with useful insight as the input synthetic RGB is intact. In principle, it should be possible to automatically collect the groundtruth dense annotation of all the objects in the game.
\subsection{Real-world Datasets}
\label{sec:camvid}
\begin{figure}
    \centering
    \includegraphics[width=0.23\textwidth]{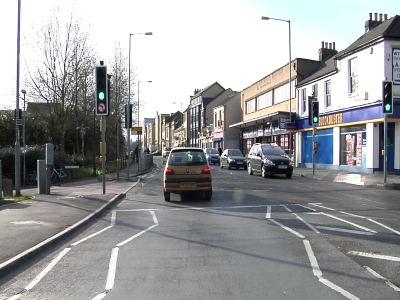}
    \includegraphics[width=0.23\textwidth]{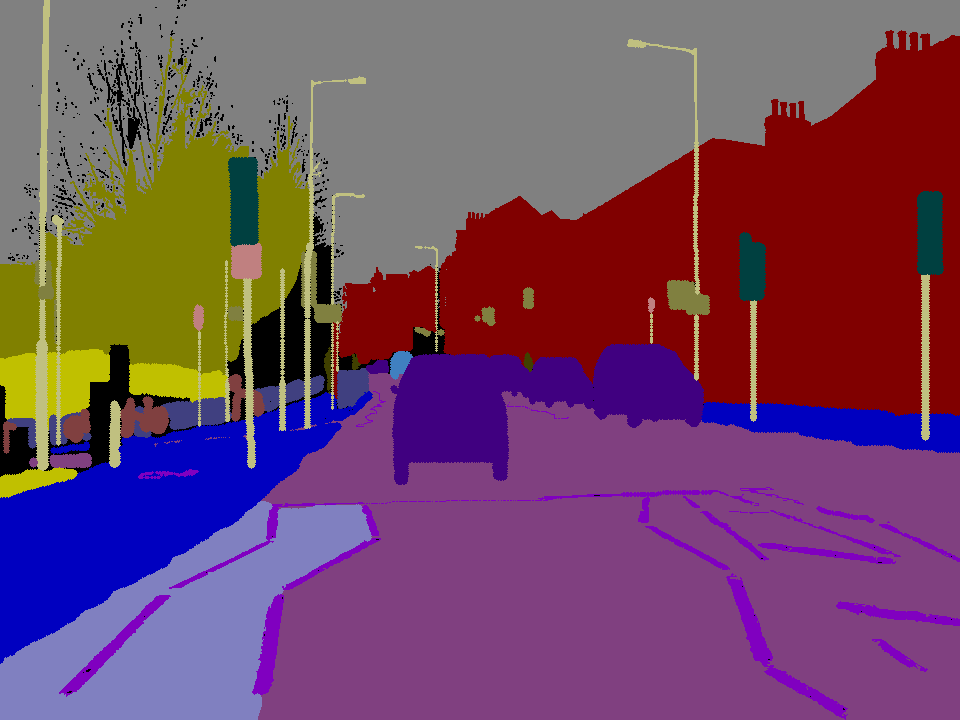}
    \includegraphics[width=0.23\textwidth]{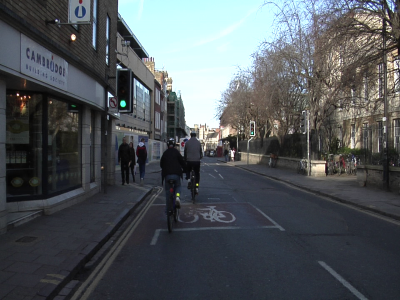}
    \includegraphics[width=0.23\textwidth]{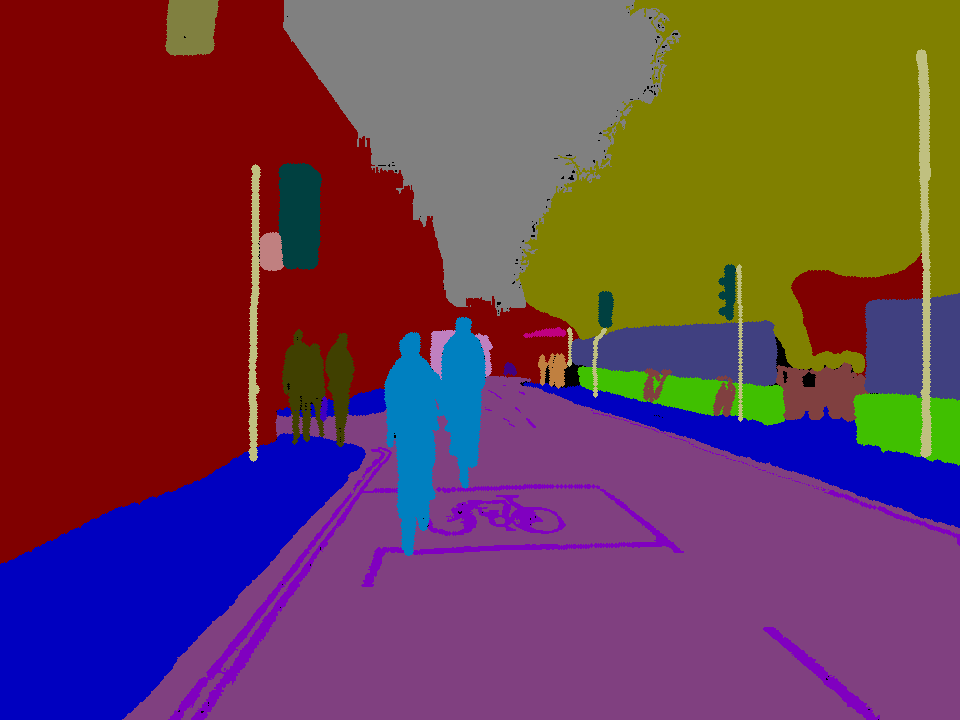}
    \caption{Two random images and the corresponding annotations from the CamVid~\cite{Brostow2009} dataset.}
    \label{fig:sample_camvid}
\end{figure}

\textbf{CamVid~\cite{Brostow2009}}. The CamVid dataset contains 701 densely labelled images collected by a driving car in a city (see Fig. \ref{fig:sample_camvid}). The label set is $\{\textrm{Sky},\,\textrm{Building},\,\textrm{Pole},\,\textrm{Road Marking},\,\textrm{Road},$ $\textrm{Pavement},\,\textrm{Tree},\,\textrm{Sign Symbol},\,\textrm{Fence},\,\textrm{Vehicle},\,\textrm{Pedestrian},\,\textrm{Bike}\}$. The data is split into 367, 101, and 233 images for train, validation, and test.

\begin{figure}
    \centering
    \includegraphics[width=0.48\textwidth]{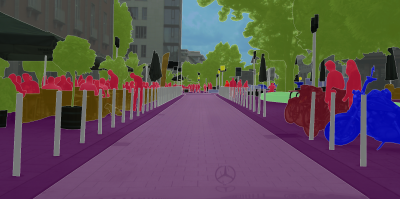}
    \includegraphics[width=0.48\textwidth]{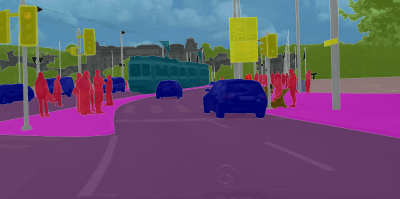}
    \caption{Two random images with the corresponding dense annotations from the Cityscapes~\cite{Cordts2016Cityscapes} dataset.}
    \label{fig:sample_cityscapes}
\end{figure}

\textbf{Cityscapes~\cite{Cordts2016Cityscapes}}. The Cityscapes dataset offers \num{5000} finely annotated images and \num{20000} coarsely annotated images of urban street scenes over 33 labels collected in cities in Germany and Switzerland. In addition to the dense and coarse image annotations, the dataset also includes car odometry readings, GPS readings, and the disparity maps that are calculated using the stereo camera. See Fig.~\ref{fig:sample_cityscapes} for pixel-level annotation examples. The densely annotated images are split into sets of 2975, 500, and 1525 for training, validation, and test. With the exception of `Road Marking', all the other 11 classes of CamVid are present in Cityscapes. The groundtruth for the test set is kept private for evaluation purposes. Thus, we perform the Cityscapes evaluations on the validation set.

\section{Dense Image Classification}
For this task we use the fully convolutional network (FCN) of Long \etal~\cite{long2015fully}. More specifically, we use the \texttt{FCN8} architecture on top of a 16-layer VGG Net~\cite{simonyan2014very}. \texttt{FCN8} is a simple architecture that only uses convolution and deconvolution to perform classification, and provides a competitive baseline for dense image classification. All of the networks will be trained with the same settings: SGD with momentum ($0.9$), and pre-defined step sizes of $10^{-4}$, $10^{-5}$, and $10^{-6}$ for $50$, $25$, and $5$ epochs. We use the MatConvNet~\cite{vedaldi15matconvnet} library.

To align our datasets with respect to the label space, we define two subsets of the data. The datasets \texttt{CamVid}, \texttt{Cityscapes}, and \texttt{VG} refer to a variation in which the label space is limited to $\{\textrm{Pedestrian},\,\textrm{Cars},\,\textrm{Trees},\,\textrm{Sky},\,\textrm{Background}\}$.
In this setting, the label space of the \texttt{VG} is precise and automatically extracted from the game.
The second variation \texttt{CamVid+}, \texttt{Cityscapes+}, and \texttt{VG+} refers to the setting in which the label space is the full 12 classes of the original CamVid dataset and the labels in the synthetic \texttt{VG+} is noisy. When we are evaluating on \texttt{Cityscapes+} we omit the missing `Road Marking' class.

\subsection{Evaluation with Fine-tuning}

\begin{figure}
    \centering
    \includegraphics[width=0.95\textwidth,trim=0.8cm 0.3cm 0cm 0cm, clip=true]{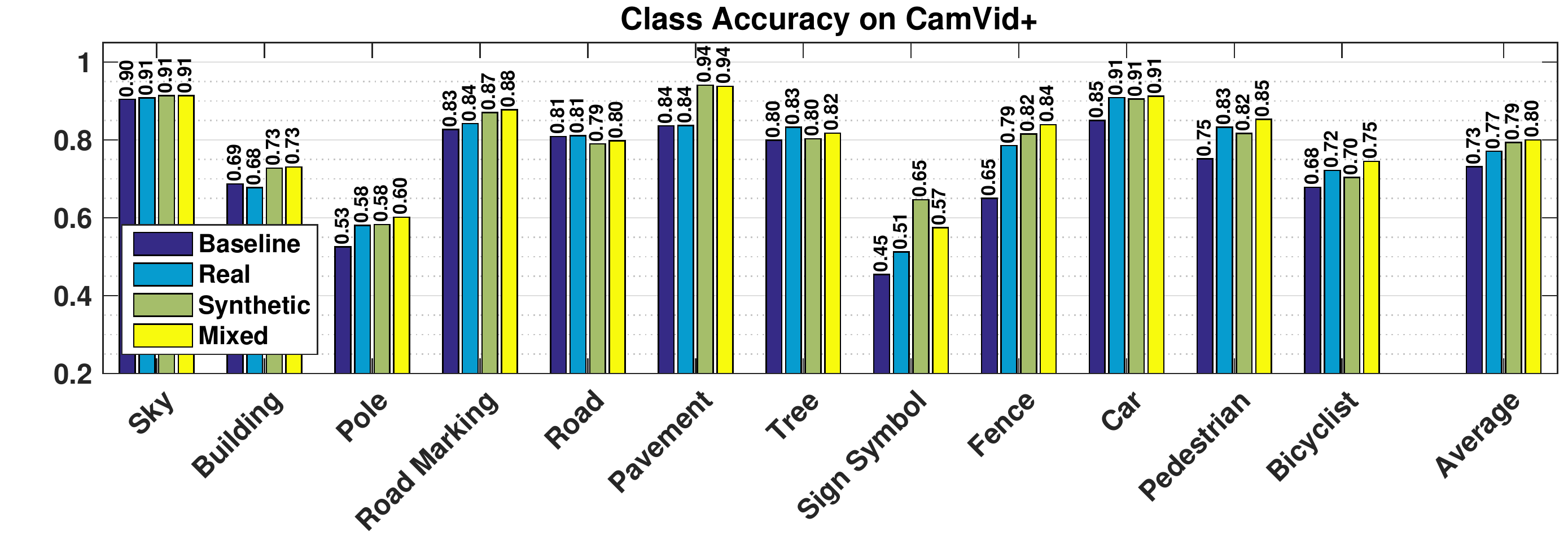}
    \caption{The per-class accuracy on the test set of \texttt{CamVid+} dataset. The \texttt{baseline} is trained on the target dataset, the \texttt{real} is pre-trained on the real alternative dataset, and the \texttt{synthetic} is pre-trained on \texttt{VG+}. The \texttt{Mixed} approach is pre-trained on both synthetic and the real alternative dataset. Pre-training the baseline with the synthetic \texttt{VG+} improves the average accuracy by $6\%$, while pre-training with the real-world \texttt{Cityscapes+} improves the average by $4\%$.}
    \label{fig:camvidp_class}
\end{figure}

To measure the effect of using synthetic data, we look at the performance in a domain adaptation setting in which we perform fine-tuning on the target dataset. In this approach we successively train our dense classifier on different datasets and then examine the performance.
In the first experiment we focus on the \texttt{CamVid+} dataset. We train four different networks and evaluate their performance on \texttt{CamVid+} to analyze the influence of our synthetic data.
Each model is pre-trained on an alternative dataset(s) before training on the target dataset. For instance, the experiment with name \texttt{Synthetic} means that the network has been pre-trained on the synthetic \texttt{VG} dataset first, and then fine-tuned on the target dataset (see Fig.~\ref{fig:camvidp_class}). The \texttt{Real} counterpart is the network that is pre-trained on the alternative real-world dataset first. The \texttt{Mixed} approach is when we pre-train on both synthetic and real-world data first.

Figure~\ref{fig:camvidp_class} compares our training strategies with respect to the per-class accuracy. Pre-training on \texttt{VG+} improves the average accuracy more than pre-training on \texttt{Cityscapes+} does, $79\%$ vs. $77\%$. The most improvement is for the class `Sign Symbol' where pre-training with \texttt{VG+} improves the accuracy by $20\%$. The highest improvement is achieved when we pre-train on both real and synthetic datasets. Table~\ref{tab:quant_vals} shows a summary of our results.

\begin{table}[]
    \centering
    \normalsize
    \renewcommand{\arraystretch}{0.9}
    \begin{tabular}{@{}lcccccc@{}}
        \toprule[1pt]
        & \multicolumn{3}{c}{CamVid+} & \multicolumn{3}{c}{Cityscapes+} \\
        \cmidrule[0.5pt](rl){2-4} \cmidrule[0.5pt](rl){5-7}
        Model   & Pixel Acc. & Class Acc. & Mean IoU    & Pixel Acc. & Class Acc. & Mean IoU\\
        \midrule
         Baseline     & 79\%   & 73\% & 47\%            & 83\%  & 77\% & 50\%\\
         \addlinespace[0.2em]         
         Real         & 80\%   & 77\% & 51\%            & 83\%   & 77\% & 50\%\\
         Synthetic & \textbf{82\%} & 79\% & 52\% & \textbf{84\%}   & \textbf{79\%} & 51\%\\
         \addlinespace[0.2em]
         Mixed        & \textbf{82\%}   & \textbf{80\%} & \textbf{53\%}   & \textbf{84\%}   & \textbf{79\%} & \textbf{52\%} \\
        \bottomrule[1pt]
    \end{tabular}
    \caption{Evaluation of different pre-training strategies on \texttt{CamVid+} and \texttt{Cityscapes+}, comparing pixel accuracy, mean class accuracy, and mean intersection over union (IoU). Pre-training on synthetic data consistently outperforms the equivalent model that is only pre-trained on real-world data. The mixed pre-training strategy gives the most improvement.}
    \label{tab:quant_vals}
\end{table}

\begin{figure}
    \centering
    \includegraphics[width=0.48\textwidth]{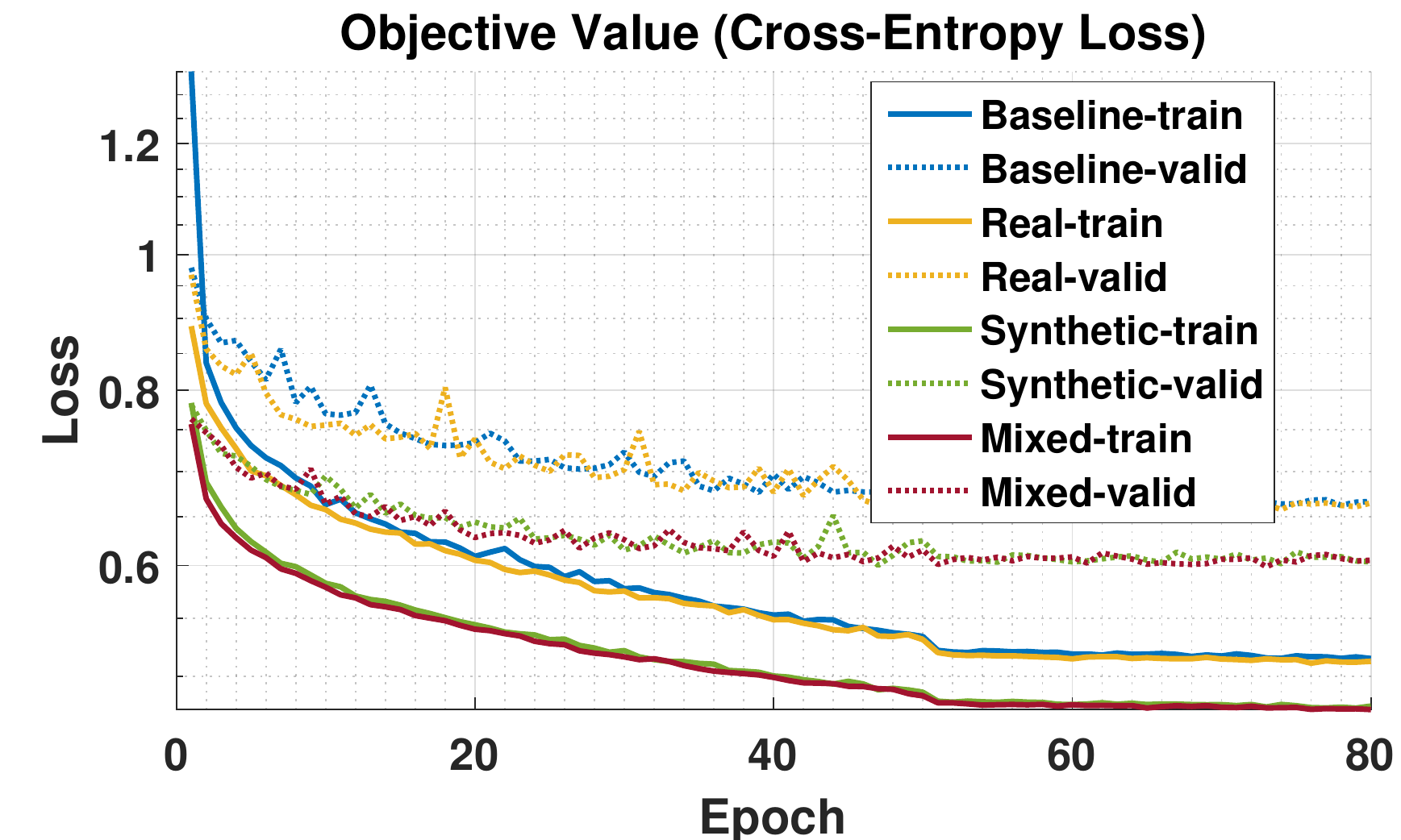}
    \includegraphics[width=0.48\textwidth]{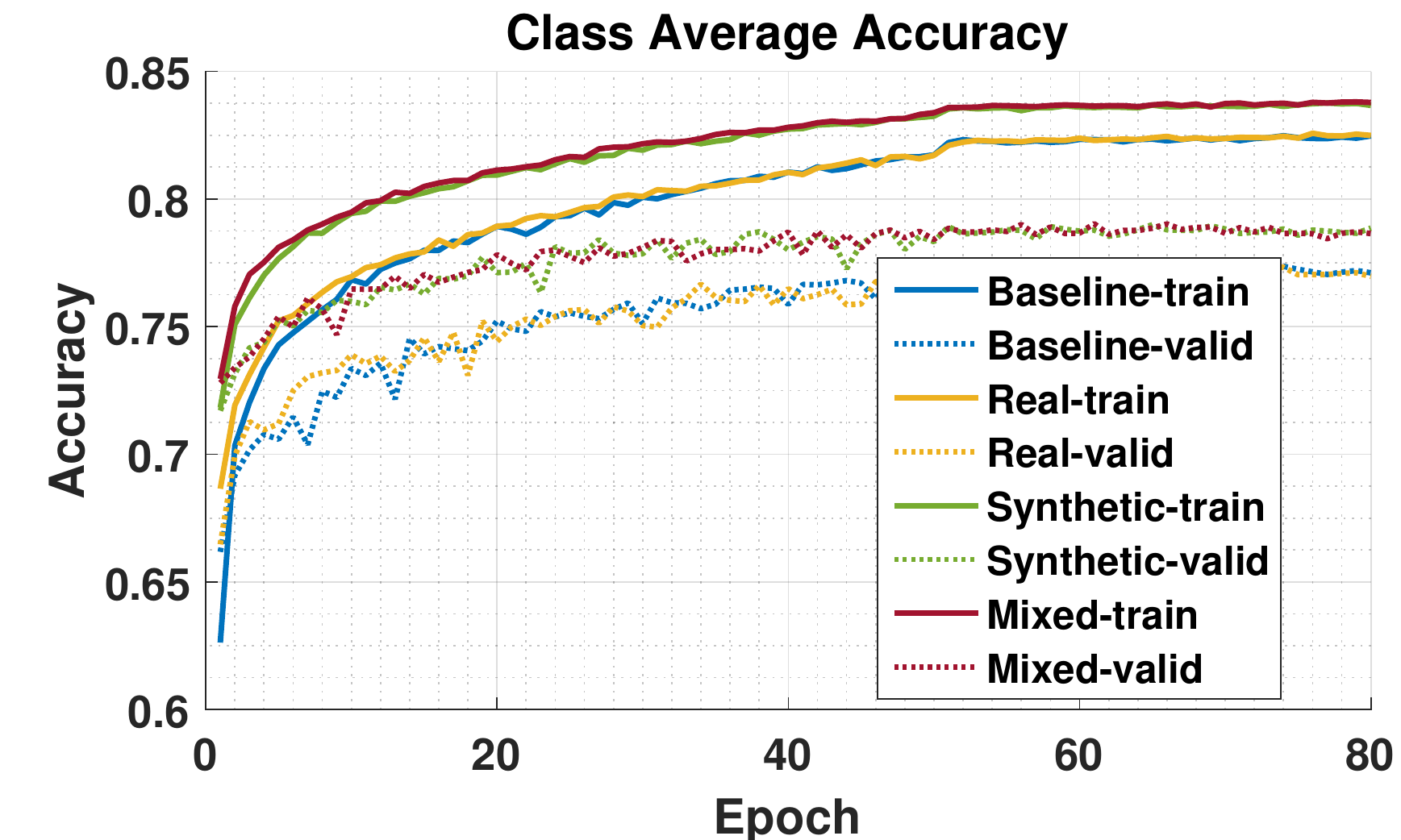}
    \caption{The influence of various pre-training approaches on the \texttt{Cityscapes+}. The left image is the objective function throughout training, and the right image is the class average accuracy. The solid lines are the evaluation results on the training set, the dashed lines are the results on the validation set. Pre-training on synthetic data gives a better initialization and final local minima compared to pre-training on a real-world dataset. The figure is best viewed in color.}
    \label{fig:eval_opt2}
\end{figure}

We can also analyze the benefits of using this synthetic data from an optimization perspective. Figure~\ref{fig:eval_opt2} shows the objective value and the class average accuracy of \texttt{Cityscapes+} during the last optimization stage. Pre-training on the \texttt{VG+} datasets yields better results on train and validation in comparison to both the baseline, and the pre-trained version on the real-world \texttt{CamVid+} dataset. Pre-training on the \texttt{VG+} data provides a better initialization and final local minima in the training procedure. In both of the results, the synthetic data provides a clear improvement over the baseline, and helps as much or more than pre-training on the real data. More figures and analysis of these experiments can be found in the supplementary material.

\subsection{Cross-dataset Evaluation}
We also evaluate the models in a cross-dataset setting in which the network is evaluated on a dataset other than the one it was trained on. We train three dense image classifiers on each dataset and examine the cross-dataset accuracy of these classifiers. The purpose of this experiment is to see how much the domain of our synthetic data differs from a real-world dataset in comparison to another real-world dataset.

\begin{figure}
\centering
    \includegraphics[width=0.48\textwidth]{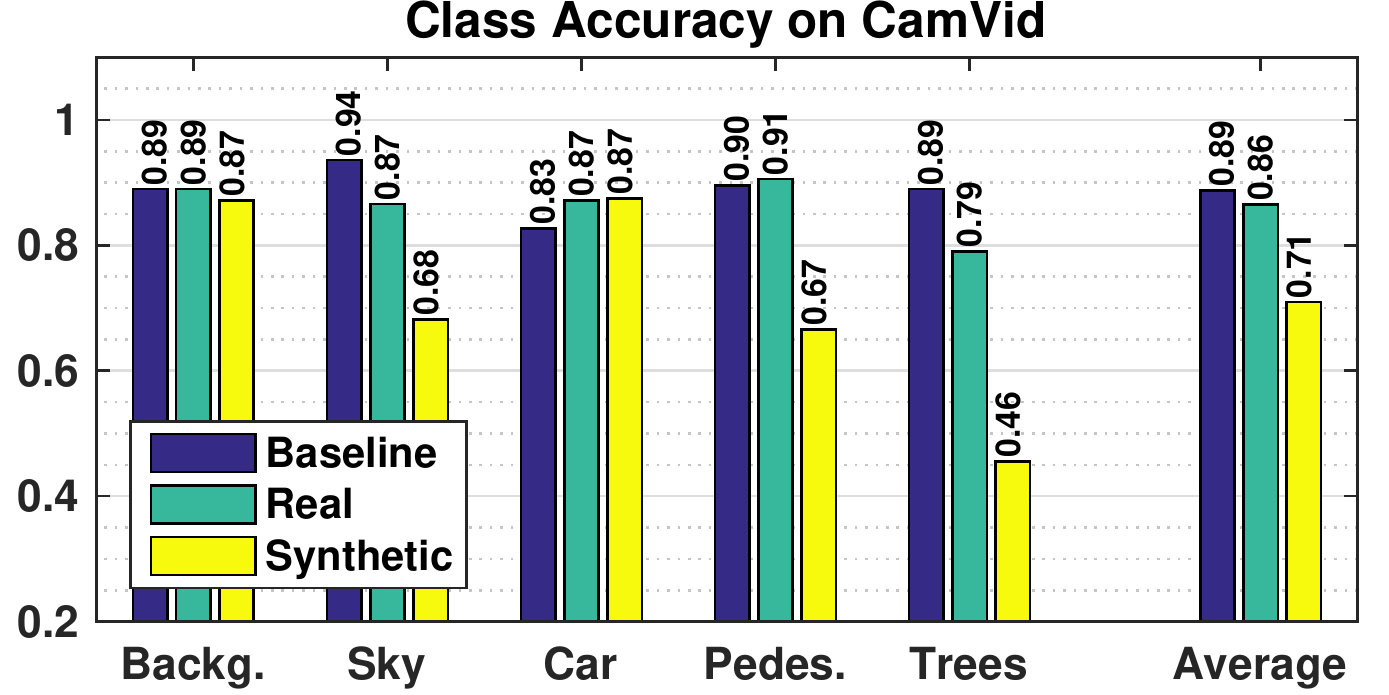}
    \includegraphics[width=0.48\textwidth]{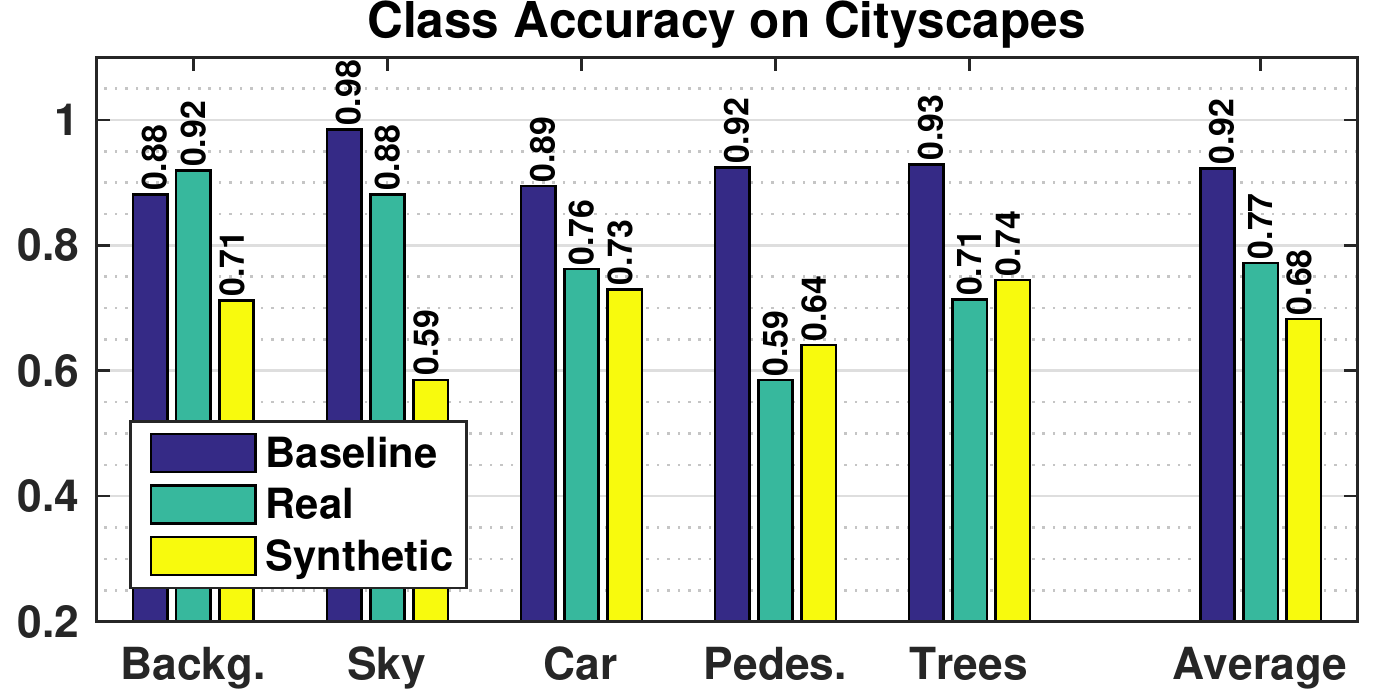}
    \caption{The cross-dataset per-class accuracy on \texttt{CamVid} (left) and \texttt{Cityscapes} (right). The baseline is trained on the target dataset, the real is trained on the real alternative dataset, and the synthetic is trained on \texttt{VG}.}
    \label{fig:cross_cam_ci_class}
\end{figure}

Figure~\ref{fig:cross_cam_ci_class} shows the cross-dataset accuracy on \texttt{CamVid} and \texttt{Cityscapes}. The first observation, as anticipated, is that the domain of a real-world dataset is more similar to the domain of another real-world dataset on average. Even though the \texttt{VG} network has been only trained on synthetic data, in `pedestrian', `car', or `trees', it competes or even in two cases outperforms the network trained on real data. Although the \texttt{VG} network does not outperform the real counterpart on average, the small gap indicates that the network with synthetic data has learned relevant features and is not overfitting to the game specific textures that can be an obstacle to generalization to the real-world domain.

\begin{figure}
    \centering
    \includegraphics[width=0.95\textwidth, trim=0.8cm 0.3cm 0cm 0cm, clip=true]{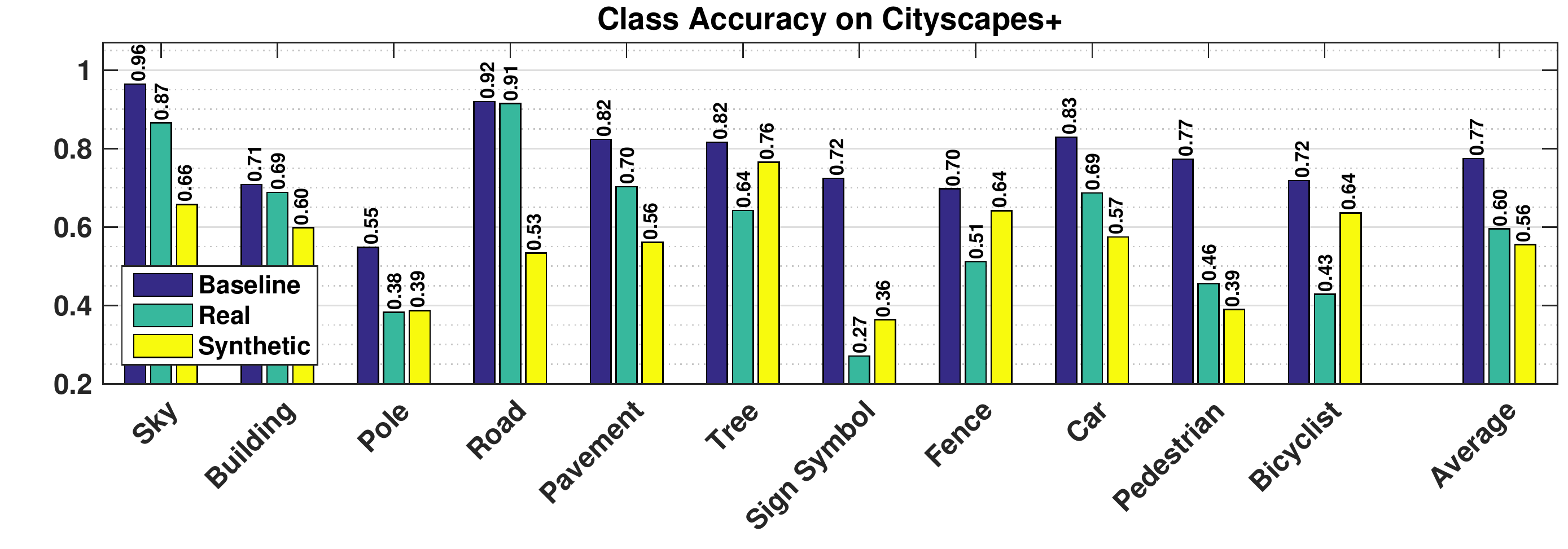}
    \caption{The cross-dataset per-class accuracy on the validation set of the \texttt{Cityscapes+} dataset. The baseline is trained on \texttt{Cityscapes+}, the real is trained on the \texttt{CamVid+}, and the synthetic is trained on \texttt{VG+}.}
    \label{fig:cityp_cross_class}
\end{figure}

Figure~\ref{fig:cityp_cross_class} shows the per-class accuracy for each network on \texttt{Cityscapes+}. Similar to the previous results, the domain of \texttt{CamVid+} is more similar to the \texttt{Cityscapes+} than the synthetic \texttt{VG+} dataset is. However, \texttt{VG+} gives better results for pole, tree, sign symbol, fence, and bicyclist. On average, the network that is trained on \texttt{CamVid+} gives a $60\%$ accuracy, and the network obtained from synthetic data gives a similar $56\%$ accuracy. The similarity in performance suggests that for the training of computer vision models the synthetic data provides a reasonable proxy to the real world images. For more figures on this experiment please see the supplementary material.

\section{Depth Estimation from RGB}
The Cityscapes dataset also provides the disparity images which we will be using in this setting. For this problem we use the method of Zoran~\etal~\cite{Zoran_2015_ICCV}, in which the underlying deep network is queried with two points in the image and has to decide which one is closer to the camera.
This method is more attractive than the other techniques because it only relies on the ordinal relationships and not on the measurement unit in the target dataset. This is useful for our experiments because the depth images of the video game are tailored to improve the visualization effects in the rendering pipeline and are not directly comparable to the measurement units of the real-world datasets. We use the same deep network architecture as Zoran~\etal~\cite{Zoran_2015_ICCV}. The images are first decomposed into superpixels. The center of each superpixel is a node in the graph. Adjacent superpixels are connected on this graph in a multiscale fashion. The two end-points of each edge are then used to query a deep network that classifies the relative depth of two input patches as $\{=,>,<\}$. A global ranking of the pixels is then generated based on these queries by solving a quadratic program. Depending on the target ordinal relationships we can estimate the depth, shading, or reflectance. We apply this method for depth estimation and train the underlying deep network on the Cityscapes and the VG datasets. While the networks in the previous experiments focused on higher level visual cues and abstractions, the nature of the network in this problem is concerned with the mid-level visual cues, which provides further insight to the quality of the synthetic data.

\begin{figure}
    \setlength{\belowcaptionskip}{-15pt}
    \centering
    \includegraphics[width=0.48\textwidth, trim=0.5cm 0cm 0cm 0cm, clip=true]{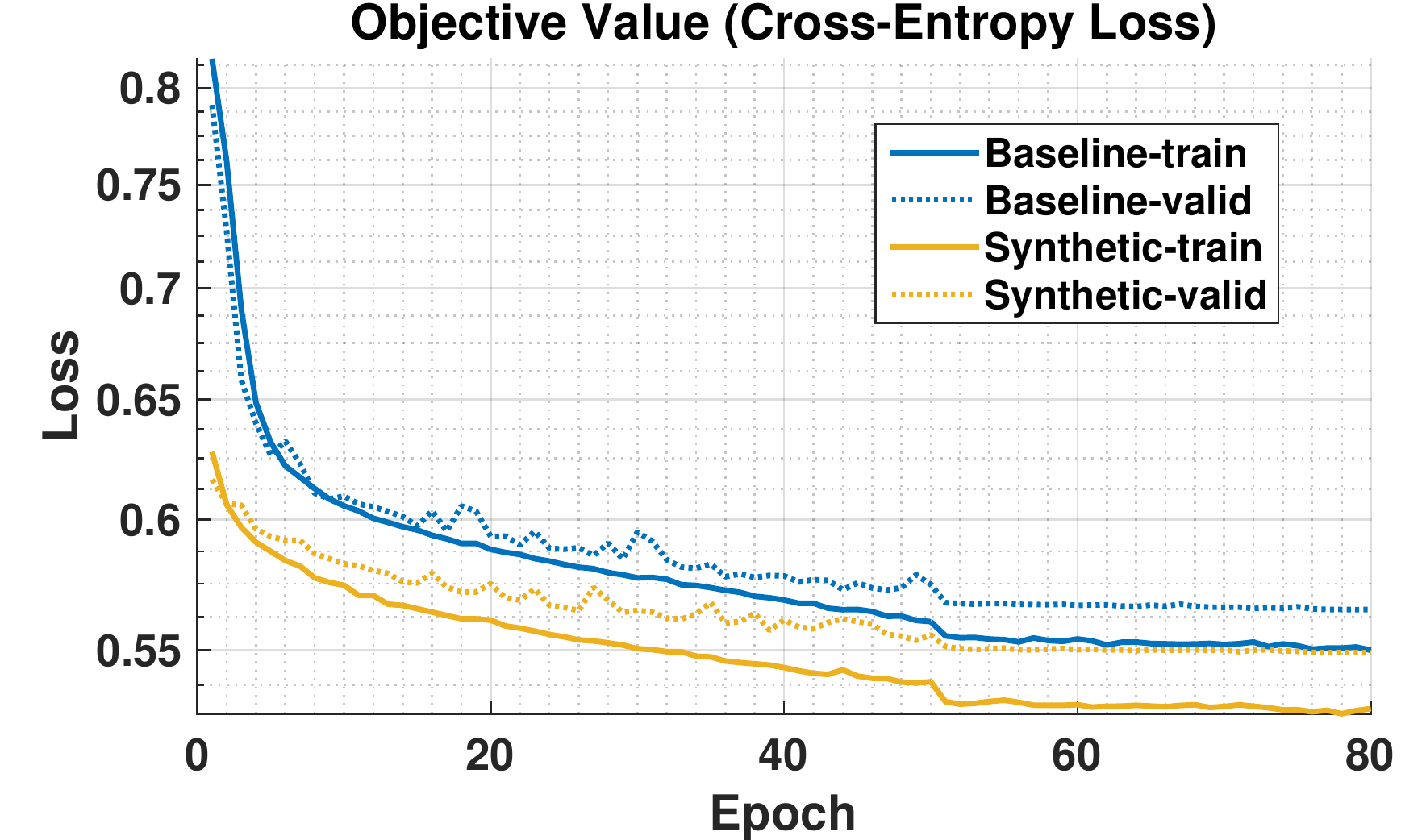}
    \includegraphics[width=0.48\textwidth, trim=0.5cm 0cm 0cm 0cm, clip=true]{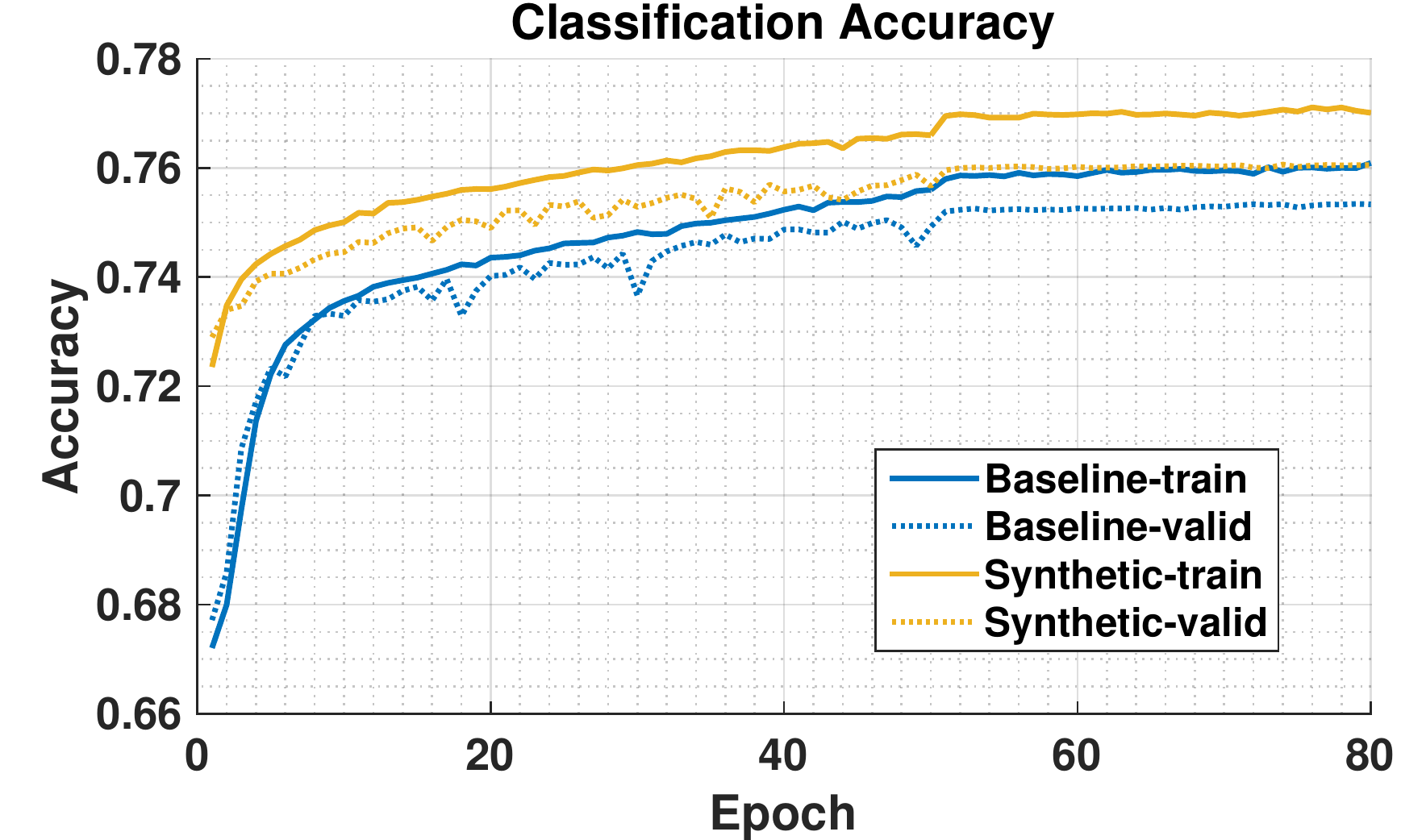}
\caption{The influence of pre-training on synthetic data for the depth estimation task on the proposed network architecture of Zoran~\etal~\cite{Zoran_2015_ICCV}.}
    \label{fig:zordinal}
\end{figure}

In Figure~\ref{fig:zordinal} we look at the influence of pre-training the network on the \texttt{VG} dataset vs. directly learning from the real dataset \texttt{Cityscapes}. Similar to the previous experiments we observe that pre-training with the synthetic data gives us a consisent improvement in initialization and the final local minima, in both validation and training. The final patch classification accuracy on the validation set is improved from 75\% to 76\%. For more figures on this experiment see the supplementary material.

\section{Conclusion}
\label{conclusion}
Concurrently, a number of independent studies that explore similar ideas have been published recently. Gaidon~\etal~\cite{Gaidon_2016_CVPR} present the Virtual KITTI dataset and show experiments on multi-object tracking tasks. Ros~\etal~\cite{Ros_2016_CVPR} present the SYNTHIA dataset of urban scenes and also demonstrate improvement in dense image segmentation using synthetic data. Our study on using video games complements the recent work by providing analysis on the photorealistic output of a state-of-the-art game engine. Furthermore, an independent study to be published by Richter~\etal~\cite{Richter_2016_ECCV} provide a similar analysis of using video games which we invite the reader to review for a complete picture.

As video games progress towards photorealistic environments, we can also use them for the training of computer vision models at no extra cost. We delivered a proof of concept by exploring the use of synthetic RGB images that we extracted from a video game. Our approach goes beyond the use of simplistic 3d CAD models as we collect a synthetic dataset that encompasses the broad context of street scenes. We presented several experiments to compare our synthetic dataset with the existing real-world ones. Our experiments show that in a cross-dataset setting, the deep neural networks that we trained on synthetic RGB images have a similar generalization power as the networks that we trained on real-world data. Furthermore, with a simple domain adaptation technique such as fine-tuning, pre-training on synthetic data consistently yielded better results than pre-training on real data. Moreover, we showed that pre-training on synthetic data resulted in a better initialization and final local minima in the optimization. On a network that classifies the ordinal relationship of image patches, we also observed how pre-training on synthetic data leads to improvement in optimization and final performance.

Our results suggest that RGB images collected from video games with photorealistic environments are potentially useful for a variety of computer vision tasks. Video games can offer an alternative way to compile large datasets for direct training or augmenting real-world datasets. Although we limited our attention to a fixed atmospheric setting to facilitate comparing to the existing real-world datasets, a key advantage is the ability to capture and use data under a variety of environmental parameters. For example, it would be easy to collect data under winter, night, fog, and other conditions where reliable real data is difficult to obtain.

\vspace{0.4cm}
\noindent
\textbf{Acknowledgments.}
We would like to thank Ankur Gupta for helpful comments and discussions. We gratefully acknowledge the support of NVIDIA Corporation through the donation of the GPUs used for this research. This work was supported in part by NSERC under Grant CRDPJ 434659-12.

\bibliography{bmvc_review}

\begin{thebibliography}{51}
\providecommand{\natexlab}[1]{#1}
\providecommand{\url}[1]{\texttt{#1}}
\expandafter\ifx\csname urlstyle\endcsname\relax
  \providecommand{\doi}[1]{doi: #1}\else
  \providecommand{\doi}{doi: \begingroup \urlstyle{rm}\Url}\fi

\bibitem[hl2()]{hl2}
{Half-Life 2}.
\newblock \url{http://www.valvesoftware.com/games/hl2.html}.

\bibitem[Achanta et~al.(2012)Achanta, Shaji, Smith, Lucchi, Fua, and
  Susstrunk]{achanta2012slic}
Radhakrishna Achanta, Appu Shaji, Kevin Smith, Aurelien Lucchi, Pascal Fua, and
  Sabine Susstrunk.
\newblock {SLIC superpixels compared to state-of-the-art superpixel methods}.
\newblock \emph{TPAMI}, 2012.

\bibitem[Aubry and Russell(2015)]{aubry2015understanding}
Mathieu Aubry and Bryan~C. Russell.
\newblock {Understanding deep features with computer-generated imagery}.
\newblock In \emph{ICCV}, 2015.

\bibitem[Aubry et~al.(2014)Aubry, Maturana, Efros, Russell, and
  Sivic]{Aubry_2014_CVPR}
Mathieu Aubry, Daniel Maturana, Alexei~A. Efros, Bryan~C. Russell, and Josef
  Sivic.
\newblock Seeing 3d chairs: Exemplar part-based 2d-3d alignment using a large
  dataset of cad models.
\newblock In \emph{CVPR}, 2014.

\bibitem[Brostow et~al.(2009)Brostow, Fauqueur, and Cipolla]{Brostow2009}
Gabriel~J. Brostow, Julien Fauqueur, and Roberto Cipolla.
\newblock {Semantic object classes in video: A high-definition ground truth
  database}.
\newblock \emph{Pattern Recognition Letters}, 2009.

\bibitem[Butler et~al.(2012)Butler, Wulff, Stanley, and
  Black]{Butler:ECCV:2012}
Daniel~J. Butler, Jonas Wulff, Garrett~B. Stanley, and Michael~J. Black.
\newblock A naturalistic open source movie for optical flow evaluation.
\newblock In \emph{ECCV}, 2012.

\bibitem[Cordts et~al.(2016)Cordts, Omran, Ramos, Rehfeld, Enzweiler, Benenson,
  Franke, Roth, and Schiele]{Cordts2016Cityscapes}
Marius Cordts, Mohamed Omran, Sebastian Ramos, Timo Rehfeld, Markus Enzweiler,
  Rodrigo Benenson, Uwe Franke, Stefan Roth, and Bernt Schiele.
\newblock {The Cityscapes dataset for semantic urban scene understanding}.
\newblock In \emph{CVPR}, 2016.

\bibitem[Dalal and Triggs(2005)]{dalal2005histograms}
Navneet Dalal and Bill Triggs.
\newblock Histograms of oriented gradients for human detection.
\newblock In \emph{CVPR}, 2005.

\bibitem[Donahue et~al.(2014)Donahue, Jia, Vinyals, Hoffman, Zhang, Tzeng, and
  Darrell]{donahue2014decaf}
Jeff Donahue, Yangqing Jia, Oriol Vinyals, Judy Hoffman, Ning Zhang, Eric
  Tzeng, and Trevor Darrell.
\newblock {DeCAF}: A deep convolutional activation feature for generic visual
  recognition.
\newblock In \emph{ICML}, 2014.

\bibitem[Eigen and Fergus(2015)]{EigenICCV15}
David Eigen and Rob Fergus.
\newblock {Predicting depth, surface normals and semantic labels with a common
  multi-scale convolutional architecture}.
\newblock In \emph{ICCV}, 2015.

\bibitem[Gaidon et~al.(2016)Gaidon, Wang, Cabon, and Vig]{Gaidon_2016_CVPR}
Adrien Gaidon, Qiao Wang, Yohann Cabon, and Eleonora Vig.
\newblock Virtual worlds as proxy for multi-object tracking analysis.
\newblock In \emph{CVPR}, 2016.

\bibitem[Ganin et~al.(2016)Ganin, Ustinova, Ajakan, Germain, Larochelle,
  Laviolette, Marchand, and Lempitsky]{ganin_jmlr}
Yaroslav Ganin, Evgeniya Ustinova, Hana Ajakan, Pascal Germain, Hugo
  Larochelle, Fran\c{c}ois Laviolette, Mario Marchand, and Victor Lempitsky.
\newblock Domain-adversarial training of neural networks.
\newblock \emph{JMLR}, 2016.

\bibitem[Geiger et~al.(2013)Geiger, Lenz, Stiller, and
  Urtasun]{geiger2013vision}
Andreas Geiger, Philip Lenz, Christoph Stiller, and Raquel Urtasun.
\newblock {Vision meets robotics: The KITTI dataset}.
\newblock \emph{The International Journal of Robotics Research}, 2013.

\bibitem[Gupta et~al.(2016)Gupta, Vedaldi, and Zisserman]{gupta2016synthetic}
Ankush Gupta, Andrea Vedaldi, and Andrew Zisserman.
\newblock Synthetic data for text localisation in natural images.
\newblock \emph{arXiv preprint arXiv:1604.06646}, 2016.

\bibitem[He et~al.(2016)He, Zhang, Ren, and Sun]{he2015deep}
Kaiming He, Xiangyu Zhang, Shaoqing Ren, and Jian Sun.
\newblock Deep residual learning for image recognition.
\newblock In \emph{CVPR}, 2016.

\bibitem[Hejrati and Ramanan(2014)]{hejrati2014analysis}
Mohsen Hejrati and Deva Ramanan.
\newblock {Analysis by synthesis: 3d object recognition by object
  reconstruction}.
\newblock In \emph{CVPR}, 2014.

\bibitem[Kaneva et~al.(2011)Kaneva, Torralba, and
  Freeman]{kaneva2011evaluation}
Biliana Kaneva, Antonio Torralba, and William~T. Freeman.
\newblock Evaluation of image features using a photorealistic virtual world.
\newblock In \emph{ICCV}, 2011.

\bibitem[Kendall et~al.(2015)Kendall, Badrinarayanan, and
  Cipolla]{kendall2015bayesian}
Alex Kendall, Vijay Badrinarayanan, and Roberto Cipolla.
\newblock Bayesian segnet: Model uncertainty in deep convolutional
  encoder-decoder architectures for scene understanding.
\newblock \emph{arXiv preprint arXiv:1511.02680}, 2015.

\bibitem[Krizhevsky et~al.(2012)Krizhevsky, Sutskever, and
  Hinton]{krizhevsky2012imagenet}
Alex Krizhevsky, Ilya Sutskever, and Geoffrey~E. Hinton.
\newblock {Imagenet classification with deep convolutional neural networks}.
\newblock In \emph{NIPS}, 2012.

\bibitem[Li et~al.(2015)Li, Shen, Dai, van~den Hengel, and He]{li2015depth}
Bo~Li, Chunhua Shen, Yuchao Dai, Anton van~den Hengel, and Mingyi He.
\newblock {Depth and surface normal estimation from monocular images using
  regression on deep features and hierarchical CRFs}.
\newblock In \emph{CVPR}, 2015.

\bibitem[Liebelt and Schmid(2010)]{liebelt2010multi}
Joerg Liebelt and Cordelia Schmid.
\newblock Multi-view object class detection with a 3d geometric model.
\newblock In \emph{CVPR}, 2010.

\bibitem[Lim et~al.(2013)Lim, Pirsiavash, and Torralba]{lim2013parsing}
Jasmine~J. Lim, Hamed Pirsiavash, and Antonio Torralba.
\newblock Parsing ikea objects: Fine pose estimation.
\newblock In \emph{ICCV}, 2013.

\bibitem[Lim et~al.(2014)Lim, Khosla, and Torralba]{lim2014fpm}
Joseph~J. Lim, Aditya Khosla, and Antonio Torralba.
\newblock {FPM: Fine pose parts-based model with 3d CAD models}.
\newblock In \emph{ECCV}, 2014.

\bibitem[Liu et~al.(2015{\natexlab{a}})Liu, Shen, and Lin]{Liu_2015_CVPR}
Fayao Liu, Chunhua Shen, and Guosheng Lin.
\newblock {Deep convolutional neural fields for depth estimation from a single
  image}.
\newblock In \emph{CVPR}, 2015{\natexlab{a}}.

\bibitem[Liu et~al.(2015{\natexlab{b}})Liu, Li, Luo, Loy, and
  Tang]{Liu_2015_ICCV}
Ziwei Liu, Xiaoxiao Li, Ping Luo, Chen-Change Loy, and Xiaoou Tang.
\newblock Semantic image segmentation via deep parsing network.
\newblock In \emph{ICCV}, 2015{\natexlab{b}}.

\bibitem[Long et~al.(2015)Long, Shelhamer, and Darrell]{long2015fully}
Jonathan Long, Evan Shelhamer, and Trevor Darrell.
\newblock {Fully convolutional networks for semantic segmentation}.
\newblock In \emph{CVPR}, 2015.

\bibitem[Marin et~al.(2010)Marin, V{\'a}zquez, Ger{\'o}nimo, and
  L{\'o}pez]{marin2010learning}
Javier Marin, David V{\'a}zquez, David Ger{\'o}nimo, and Antonio~M. L{\'o}pez.
\newblock Learning appearance in virtual scenarios for pedestrian detection.
\newblock In \emph{CVPR}, 2010.

\bibitem[Pan and Yang(2010)]{pan2010survey}
Sinno~Jialin Pan and Qiang Yang.
\newblock A survey on transfer learning.
\newblock \emph{IEEE Transactions on Knowledge and Data Engineering}, 2010.

\bibitem[Papandreou et~al.(2015)Papandreou, Chen, Murphy, and
  Yuille]{Papandreou2015}
George Papandreou, Liang Chen, Kevin Murphy, and Alan~L. Yuille.
\newblock Weakly- and semi-supervised learning of a deep convolutional network
  for semantic image segmentation.
\newblock In \emph{ICCV}, 2015.

\bibitem[Peng et~al.(2015)Peng, Sun, Ali, and Saenko]{Peng_2015_ICCV}
Xingchao Peng, Baochen Sun, Karim Ali, and Kate Saenko.
\newblock Learning deep object detectors from 3d models.
\newblock In \emph{ICCV}, 2015.

\bibitem[Rematas et~al.(2014)Rematas, Ritschel, Fritz, and
  Tuytelaars]{rematas2014image}
Konstantinos Rematas, Tobias Ritschel, Mario Fritz, and Tinne Tuytelaars.
\newblock Image-based synthesis and re-synthesis of viewpoints guided by 3d
  models.
\newblock In \emph{CVPR}, 2014.

\bibitem[Richter et~al.(2016)Richter, Vineet, Roth, and
  Koltun]{Richter_2016_ECCV}
Stephan~R. Richter, Vibhav Vineet, Stefan Roth, and Vladlen Koltun.
\newblock Playing for data: {G}round truth from computer games.
\newblock In \emph{ECCV}, 2016.

\bibitem[Rogez et~al.(2015)Rogez, III, and Ramanan]{rogez2015first}
Gr{\'{e}}gory Rogez, James S.~Supancic III, and Deva Ramanan.
\newblock First-person pose recognition using egocentric workspaces.
\newblock In \emph{CVPR}, 2015.

\bibitem[Ros et~al.(2016)Ros, Sellart, Materzynska, Vazquez, and
  Lopez]{Ros_2016_CVPR}
German Ros, Laura Sellart, Joanna Materzynska, David Vazquez, and Antonio~M.
  Lopez.
\newblock The {SYNTHIA} dataset: A large collection of synthetic images for
  semantic segmentation of urban scenes.
\newblock In \emph{CVPR}, 2016.

\bibitem[Russakovsky et~al.(2015)Russakovsky, Deng, Su, Krause, Satheesh, Ma,
  Huang, Karpathy, Khosla, Bernstein, Berg, and Fei-Fei]{ILSVRC15}
Olga Russakovsky, Jia Deng, Hao Su, Jonathan Krause, Sanjeev Satheesh, Sean Ma,
  Zhiheng Huang, Andrej Karpathy, Aditya Khosla, Michael Bernstein,
  Alexander~C. Berg, and Li~Fei-Fei.
\newblock Imagenet large scale visual recognition challenge.
\newblock \emph{IJCV}, 2015.

\bibitem[Saxena et~al.(2008)Saxena, Chung, and Ng]{saxena20083}
Ashutosh Saxena, Sung~H. Chung, and Andrew~Y. Ng.
\newblock {3-D depth reconstruction from a single still image}.
\newblock \emph{IJCV}, 2008.

\bibitem[Scharw{\"a}chter et~al.(2013)Scharw{\"a}chter, Enzweiler, Franke, and
  Roth]{scharwachter2013efficient}
Timo Scharw{\"a}chter, Markus Enzweiler, Uwe Franke, and Stefan Roth.
\newblock Efficient multi-cue scene segmentation.
\newblock In \emph{Pattern Recognition}. 2013.

\bibitem[Shafaei and Little(2016)]{Shafaei2016}
Alireza Shafaei and James~J. Little.
\newblock Real-time human motion capture with multiple depth cameras.
\newblock In \emph{CRV}, 2016.

\bibitem[Sharif~Razavian et~al.(2014)Sharif~Razavian, Azizpour, Sullivan, and
  Carlsson]{sharif2014cnn}
Ali Sharif~Razavian, Hossein Azizpour, Josephine Sullivan, and Stefan Carlsson.
\newblock Cnn features off-the-shelf: an astounding baseline for recognition.
\newblock In \emph{CVPR Workshops}, 2014.

\bibitem[Shotton et~al.(2013)Shotton, Girshick, Fitzgibbon, Sharp, Cook,
  Finocchio, Moore, Kohli, Criminisi, and Kipman]{shotton2013efficient}
Jamie Shotton, Ross Girshick, Andrew Fitzgibbon, Toby Sharp, Mat Cook, Mark
  Finocchio, Richard Moore, Pushmeet Kohli, Antonio Criminisi, and Alex Kipman.
\newblock {Efficient human pose estimation from single depth images}.
\newblock \emph{TPAMI}, 2013.

\bibitem[Simonyan and Zisserman(2014)]{simonyan2014very}
Karen Simonyan and Andrew Zisserman.
\newblock Very deep convolutional networks for large-scale image recognition.
\newblock \emph{arXiv preprint arXiv:1409.1556}, 2014.

\bibitem[Stark et~al.(2010)Stark, Goesele, and Schiele]{stark2010back}
Michael Stark, Michael Goesele, and Bernt Schiele.
\newblock {Back to the future: learning shape models from 3d CAD data}.
\newblock In \emph{BMVC}, 2010.

\bibitem[Sun and Saenko(2014)]{sun2014virtual}
Baochen Sun and Kate Saenko.
\newblock From virtual to reality: Fast adaptation of virtual object detectors
  to real domains.
\newblock In \emph{BMVC}, 2014.

\bibitem[Taylor et~al.(2007)Taylor, Chosak, and Brewer]{taylor2007ovvv}
Geoffrey~R. Taylor, Andrew~J. Chosak, and Paul~C. Brewer.
\newblock {OVVV: Using virtual worlds to design and evaluate surveillance
  systems}.
\newblock In \emph{CVPR}, 2007.

\bibitem[Tompson et~al.(2014)Tompson, Stein, Lecun, and
  Perlin]{tompson2014real}
Jonathan Tompson, Murphy Stein, Yann Lecun, and Ken Perlin.
\newblock {Real-time continuous pose recovery of human hands using
  convolutional networks}.
\newblock \emph{TOG}, 2014.

\bibitem[Vedaldi and Lenc(2015)]{vedaldi15matconvnet}
Andrea Vedaldi and Karel Lenc.
\newblock {MatConvNet} -- {Convolutional} neural networks for {MATLAB}.
\newblock In \emph{Proceeding of the {ACM} Int. Conf. on Multimedia}, 2015.

\bibitem[Yosinski et~al.(2014)Yosinski, Clune, Bengio, and
  Lipson]{yosinski_2014_NIPS}
Jason Yosinski, Jeff Clune, Yoshua Bengio, and Hod Lipson.
\newblock How transferable are features in deep neural networks?
\newblock In \emph{NIPS}. 2014.

\bibitem[Yu and Koltun(2016)]{YuKoltun2016}
Fisher Yu and Vladlen Koltun.
\newblock Multi-scale context aggregation by dilated convolutions.
\newblock In \emph{ICLR}, 2016.

\bibitem[Zheng et~al.(2015)Zheng, Jayasumana, Romera-Paredes, Vineet, Su, Du,
  Huang, and Torr]{crfasrnn_iccv2015}
Shuai Zheng, Sadeep Jayasumana, Bernardino Romera-Paredes, Vibhav Vineet,
  Zhizhong Su, Dalong Du, Chang Huang, and Philip Torr.
\newblock Conditional random fields as recurrent neural networks.
\newblock In \emph{ICCV}, 2015.

\bibitem[Zhuo et~al.(2015)Zhuo, Salzmann, He, and Liu]{Zhuo_2015_CVPR}
Wei Zhuo, Mathieu Salzmann, Xuming He, and Miaomiao Liu.
\newblock Indoor scene structure analysis for single image depth estimation.
\newblock In \emph{CVPR}, 2015.

\bibitem[Zoran et~al.(2015)Zoran, Isola, Krishnan, and
  Freeman]{Zoran_2015_ICCV}
Daniel Zoran, Phillip Isola, Dilip Krishnan, and William~T. Freeman.
\newblock Learning ordinal relationships for mid-level vision.
\newblock In \emph{ICCV}, 2015.

\end{thebibliography}
\newpage
\section*{Supplementary Material}
\subsection*{Evaluation with Fine-tuning}
\begin{figure}
    \includegraphics[width=0.5\textwidth]{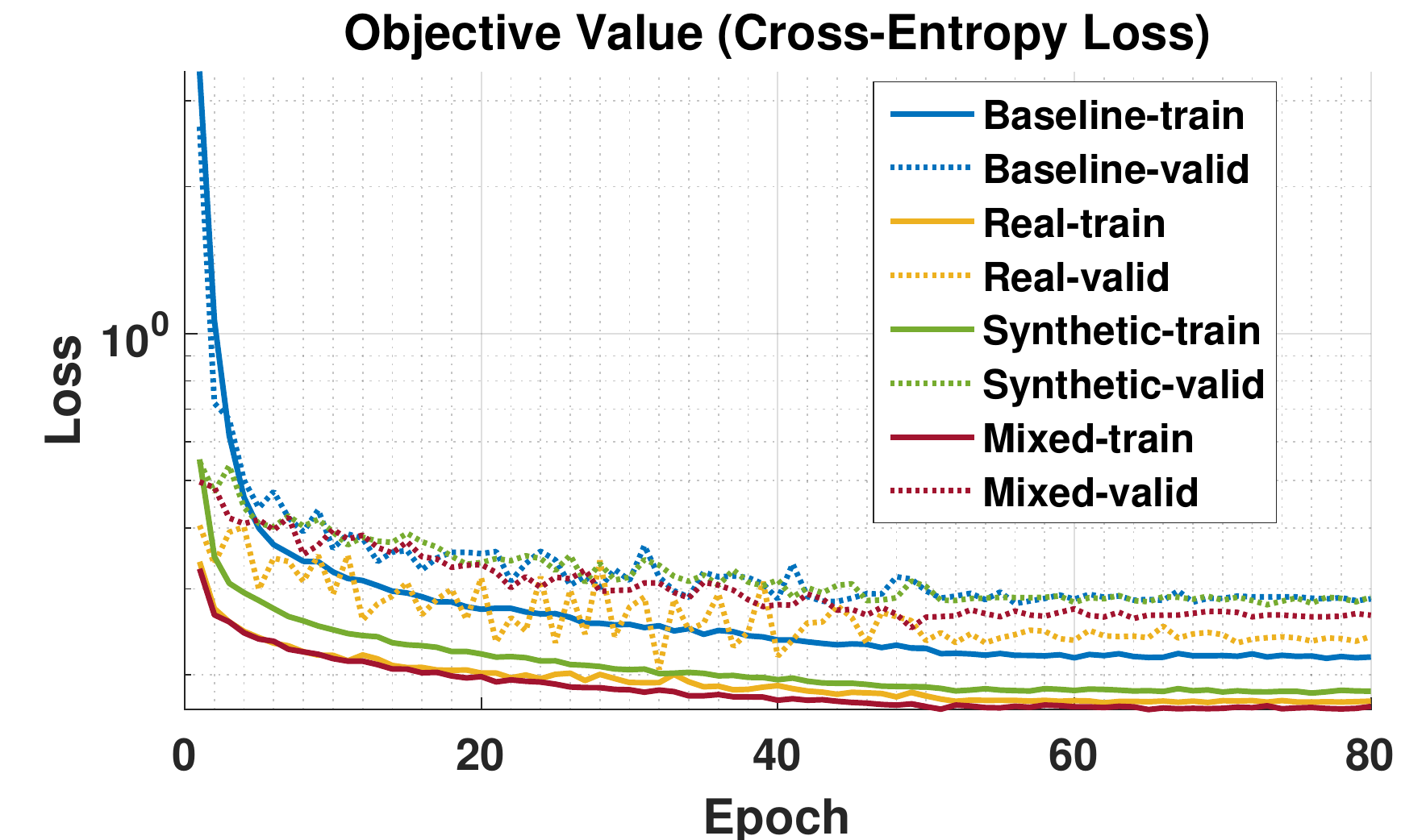}
    \includegraphics[width=0.5\textwidth]{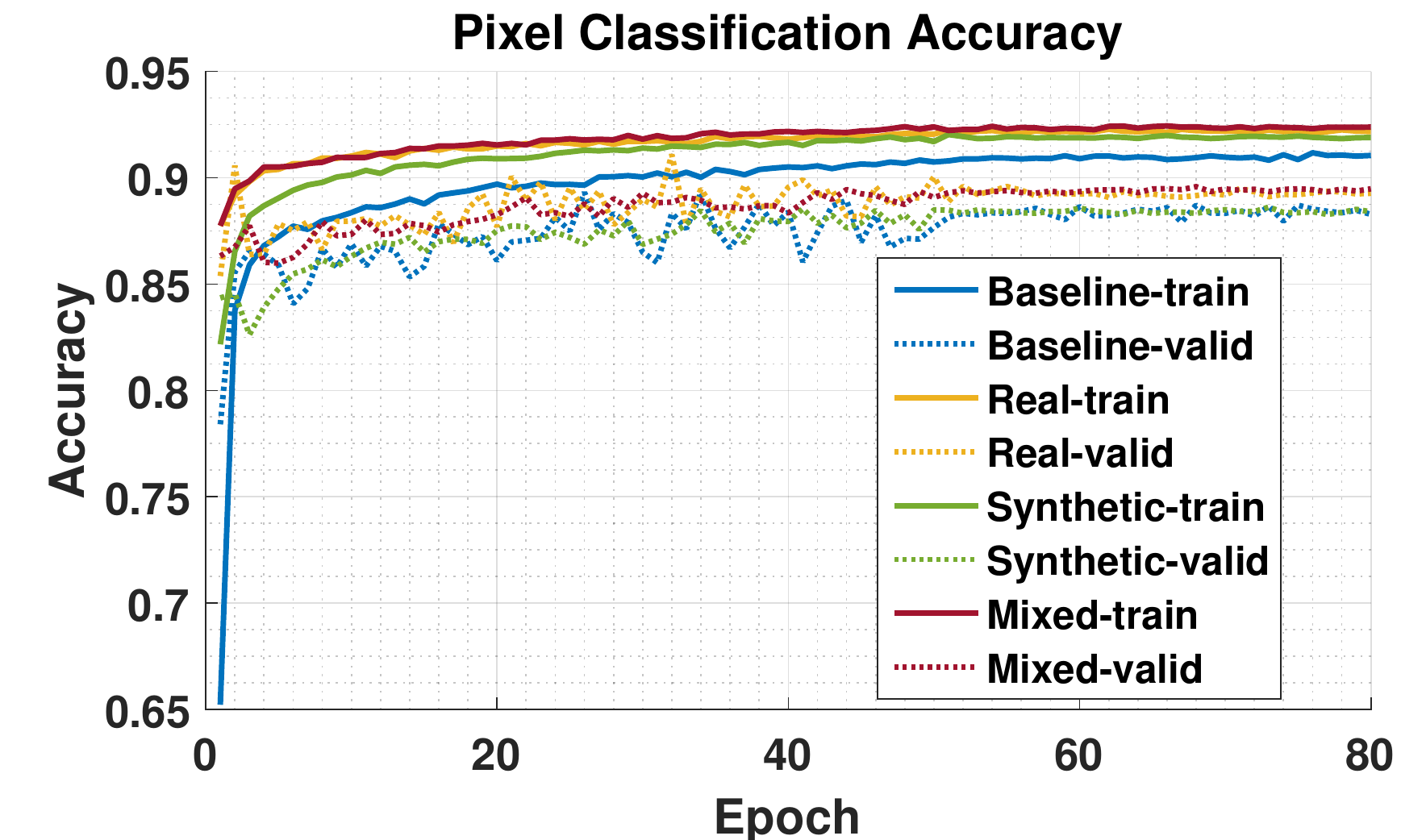}
    \includegraphics[width=0.5\textwidth]{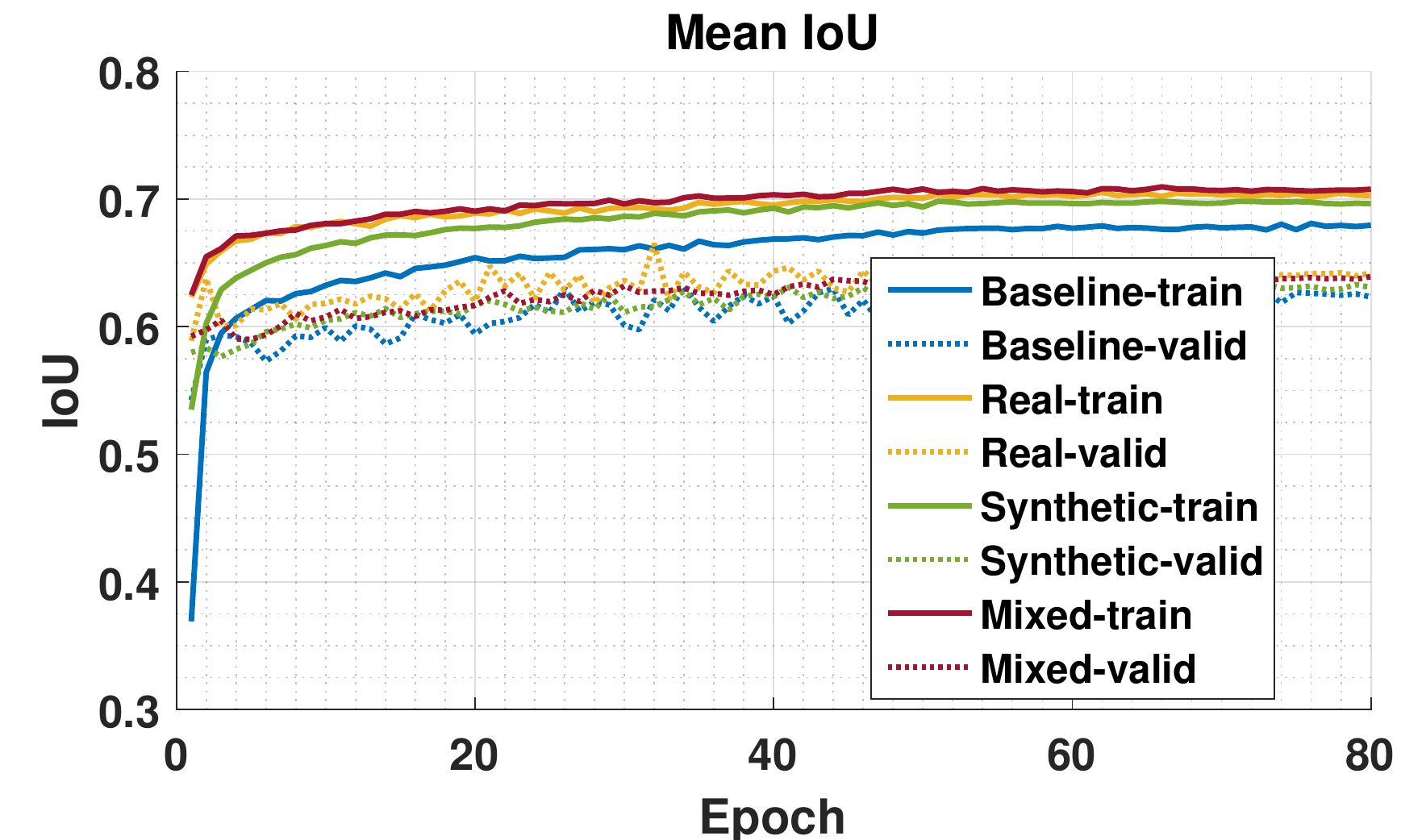}
    \includegraphics[width=0.5\textwidth]{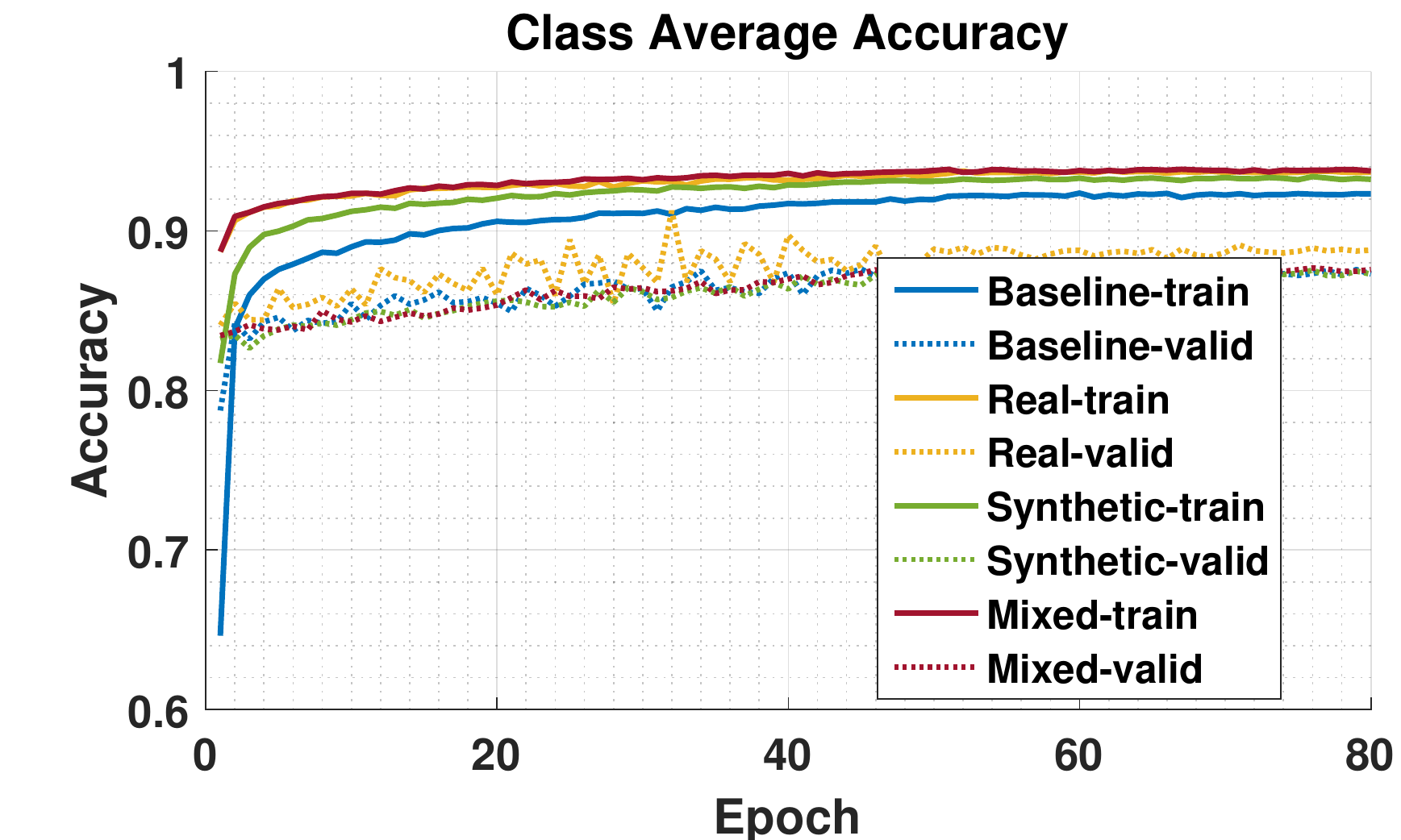}
    \caption{The influence of various pre-training approaches on the \texttt{CamVid} dataset. The solid lines are the evaluation results on the training set, the dashed lines are the results on the \textit{validation} set.}
    \label{sup:fig:camvid_conv}
\end{figure}

\begin{figure}
    \includegraphics[width=0.5\textwidth]{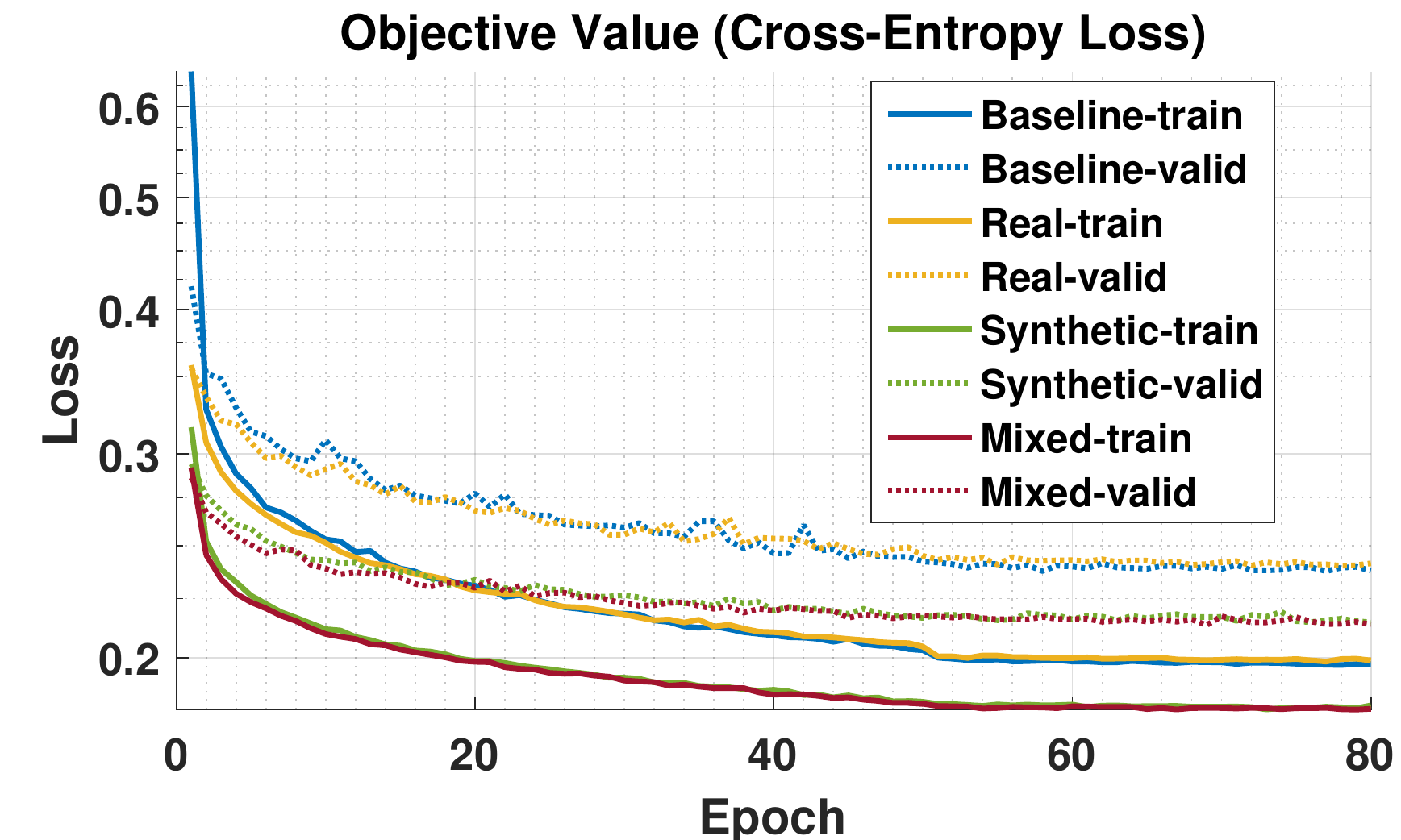}
    \includegraphics[width=0.5\textwidth]{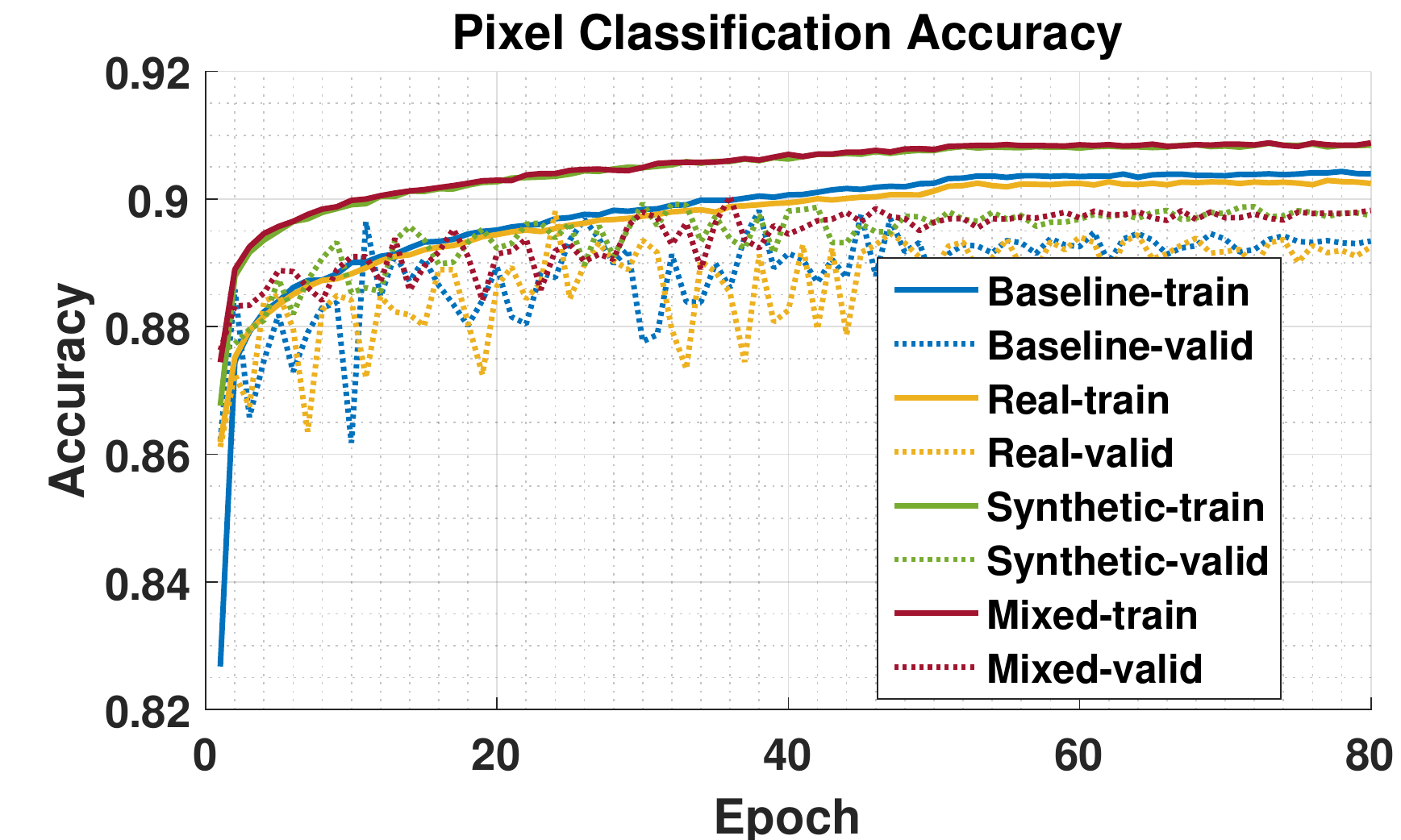}
    \includegraphics[width=0.5\textwidth]{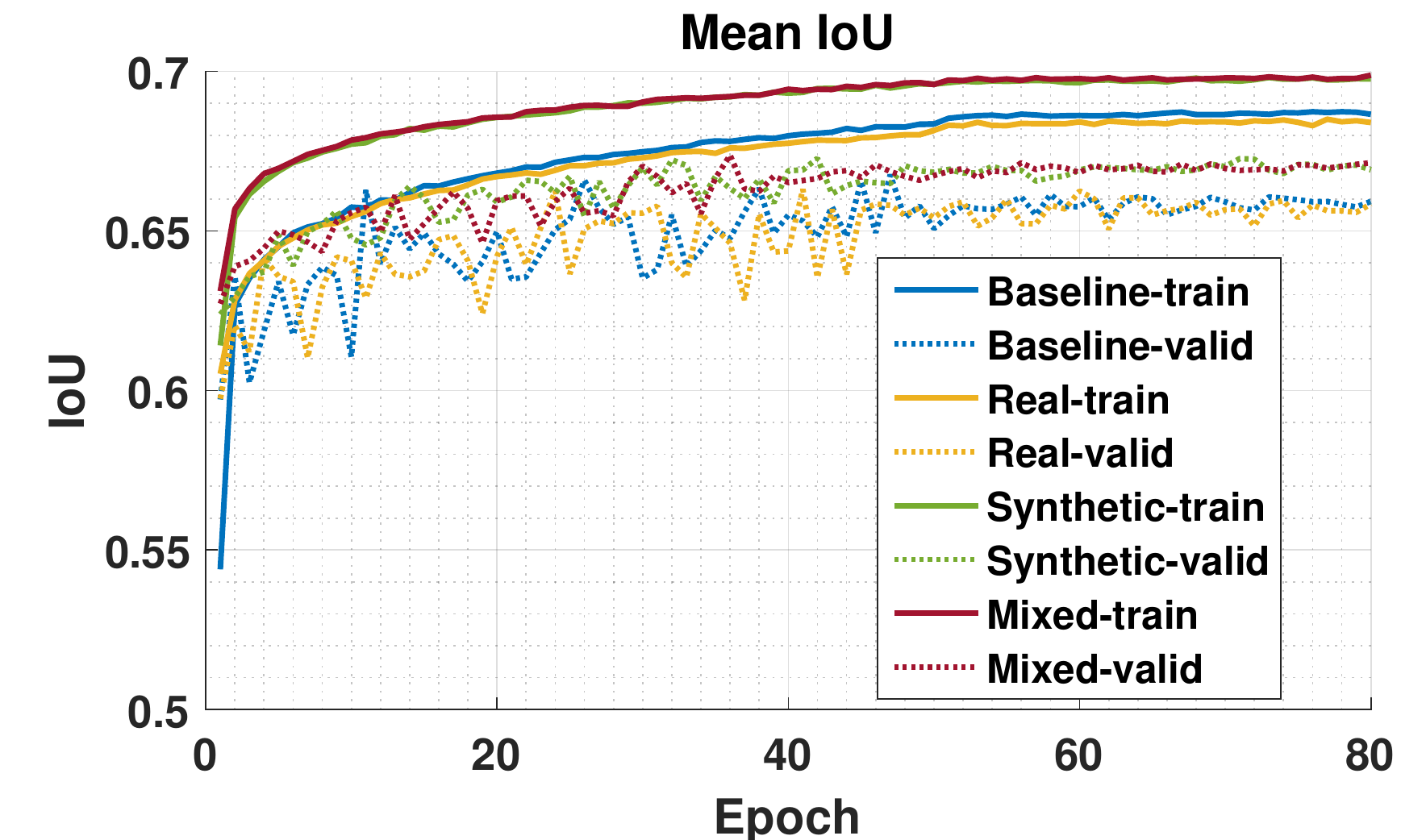}
    \includegraphics[width=0.5\textwidth]{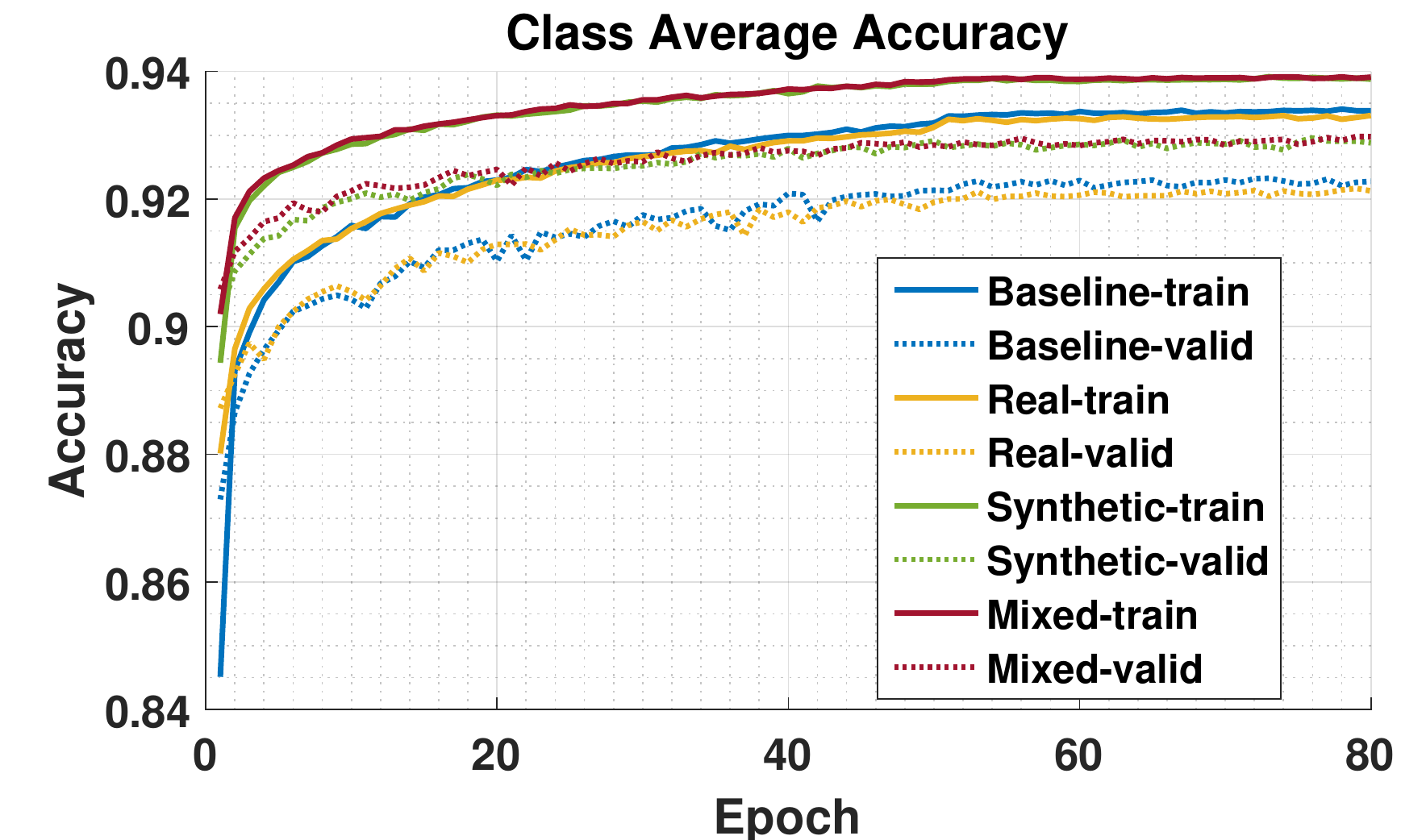}
    \caption{The influence of various pre-training approaches on the \texttt{Cityscapes} dataset. The solid lines are the evaluation results on the training set, the dashed lines are the results on the \textit{validation} set.}
    \label{sup:fig:city_conv}
\end{figure}

\begin{figure}
    \includegraphics[width=0.5\textwidth]{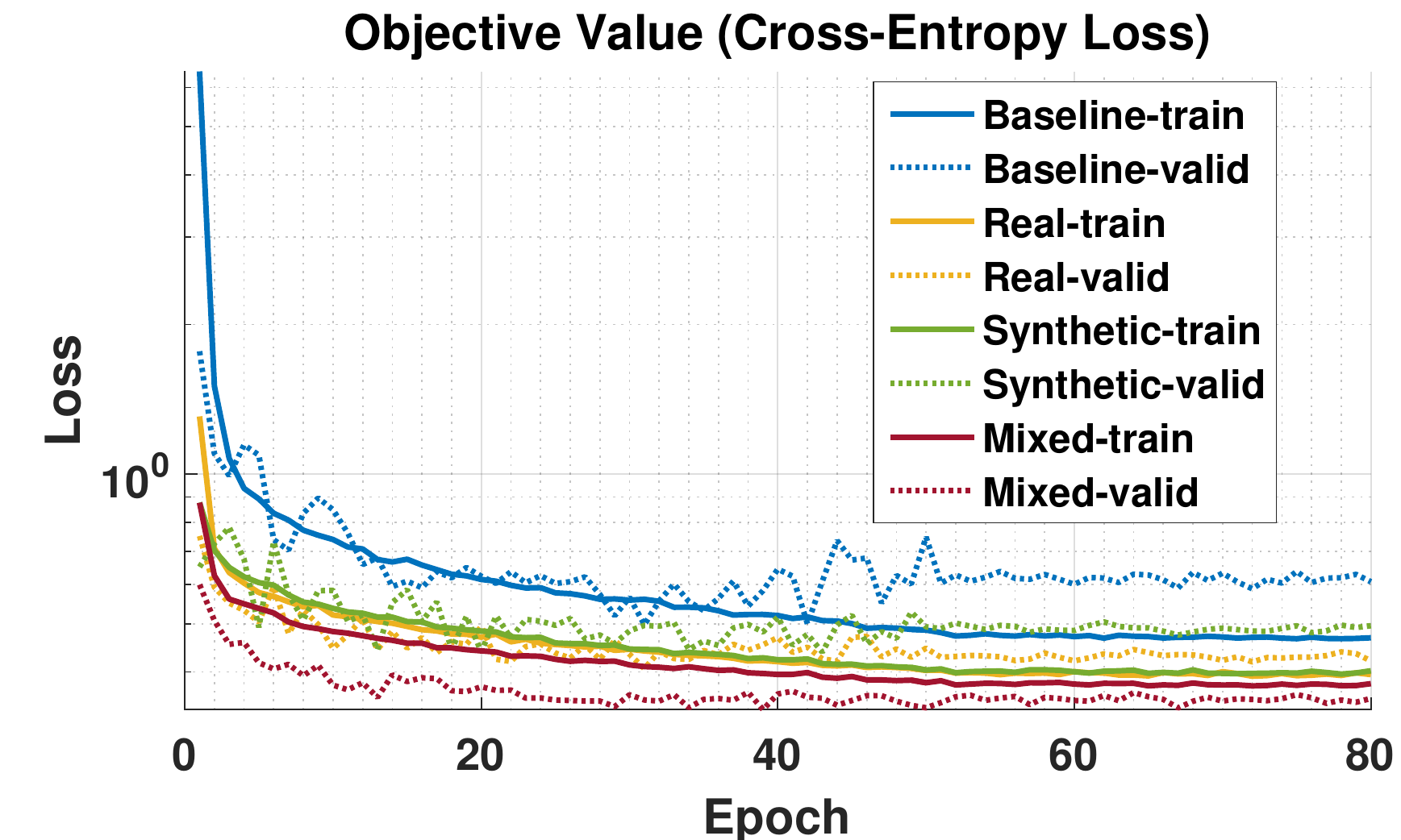}
    \includegraphics[width=0.5\textwidth]{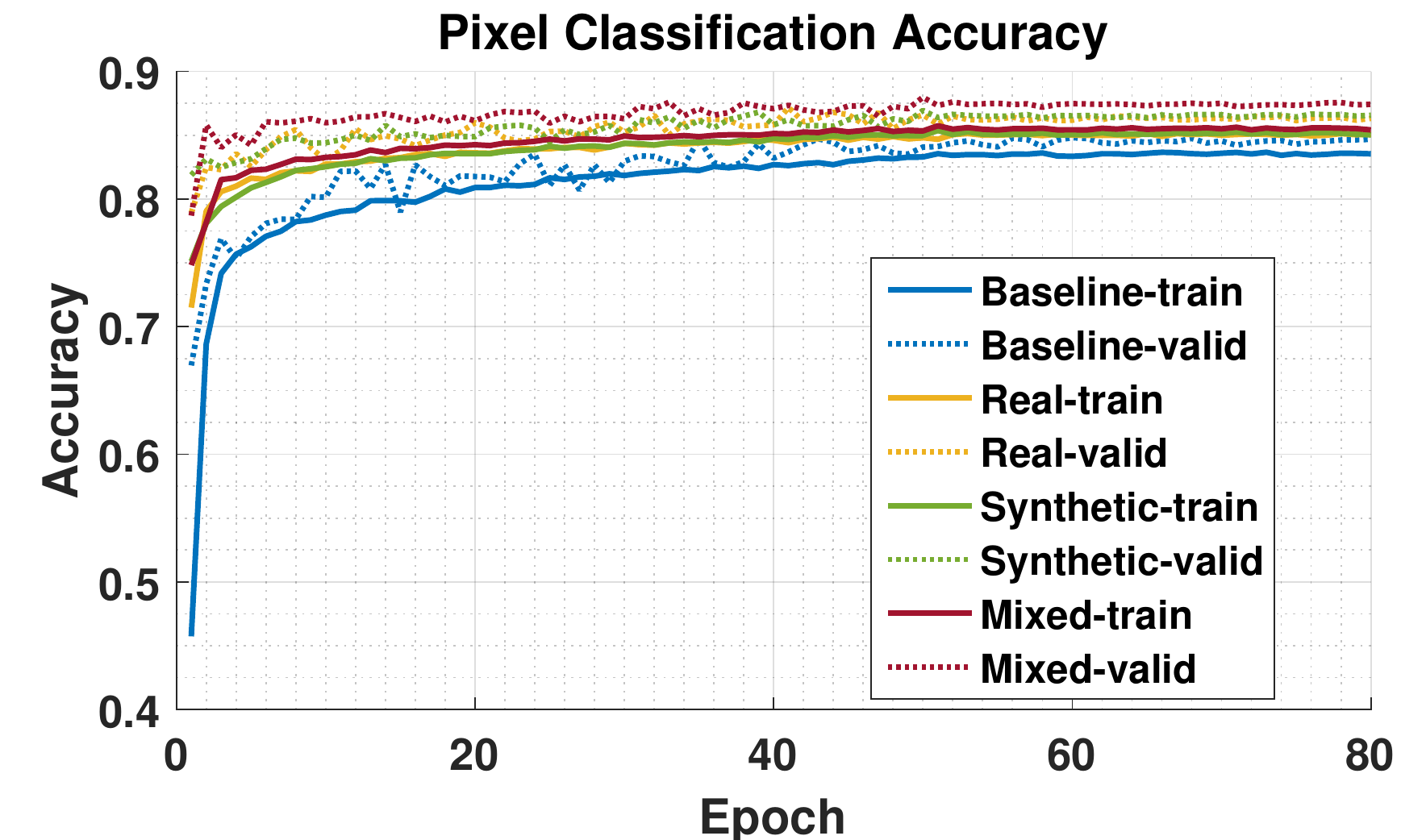}
    \includegraphics[width=0.5\textwidth]{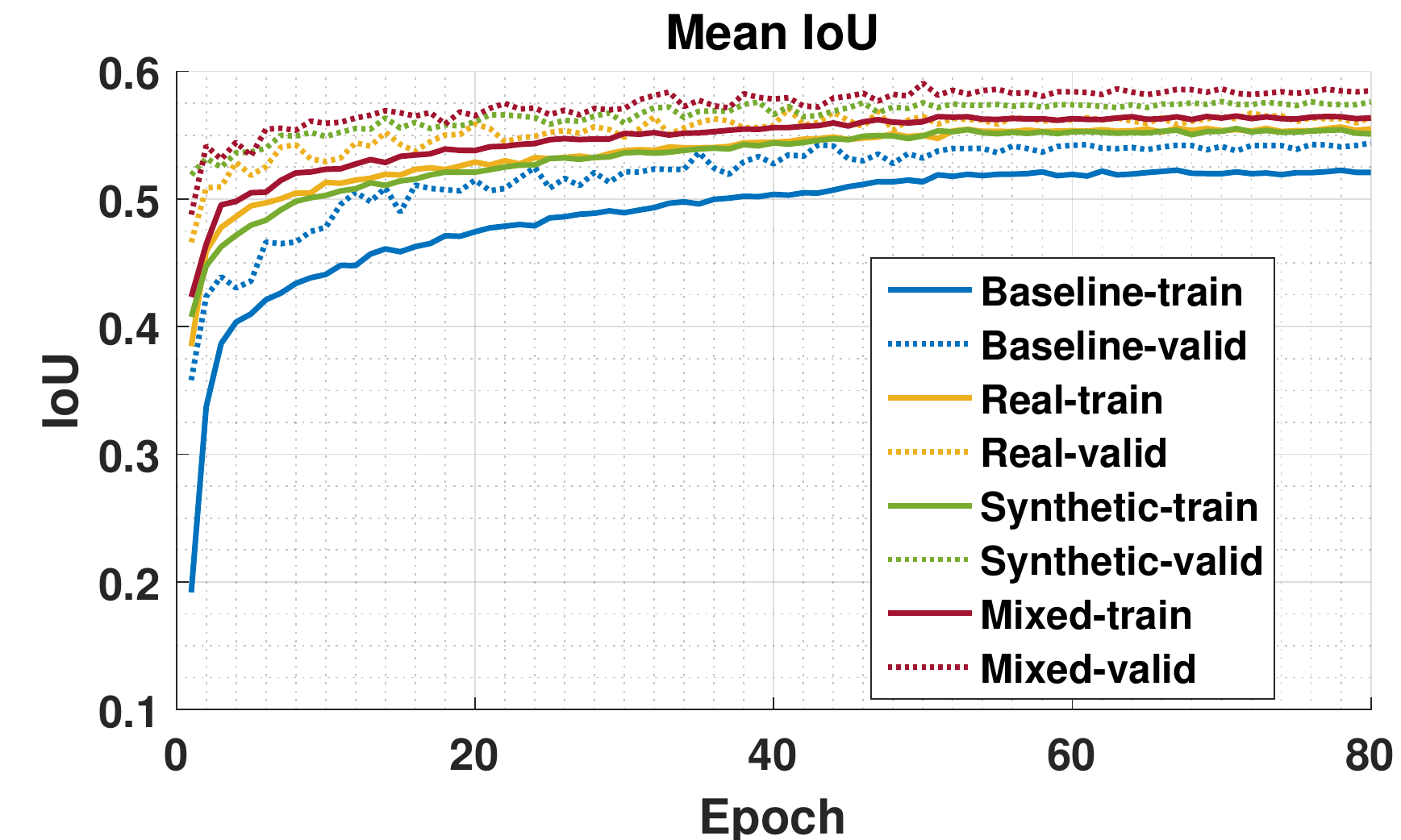}
    \includegraphics[width=0.5\textwidth]{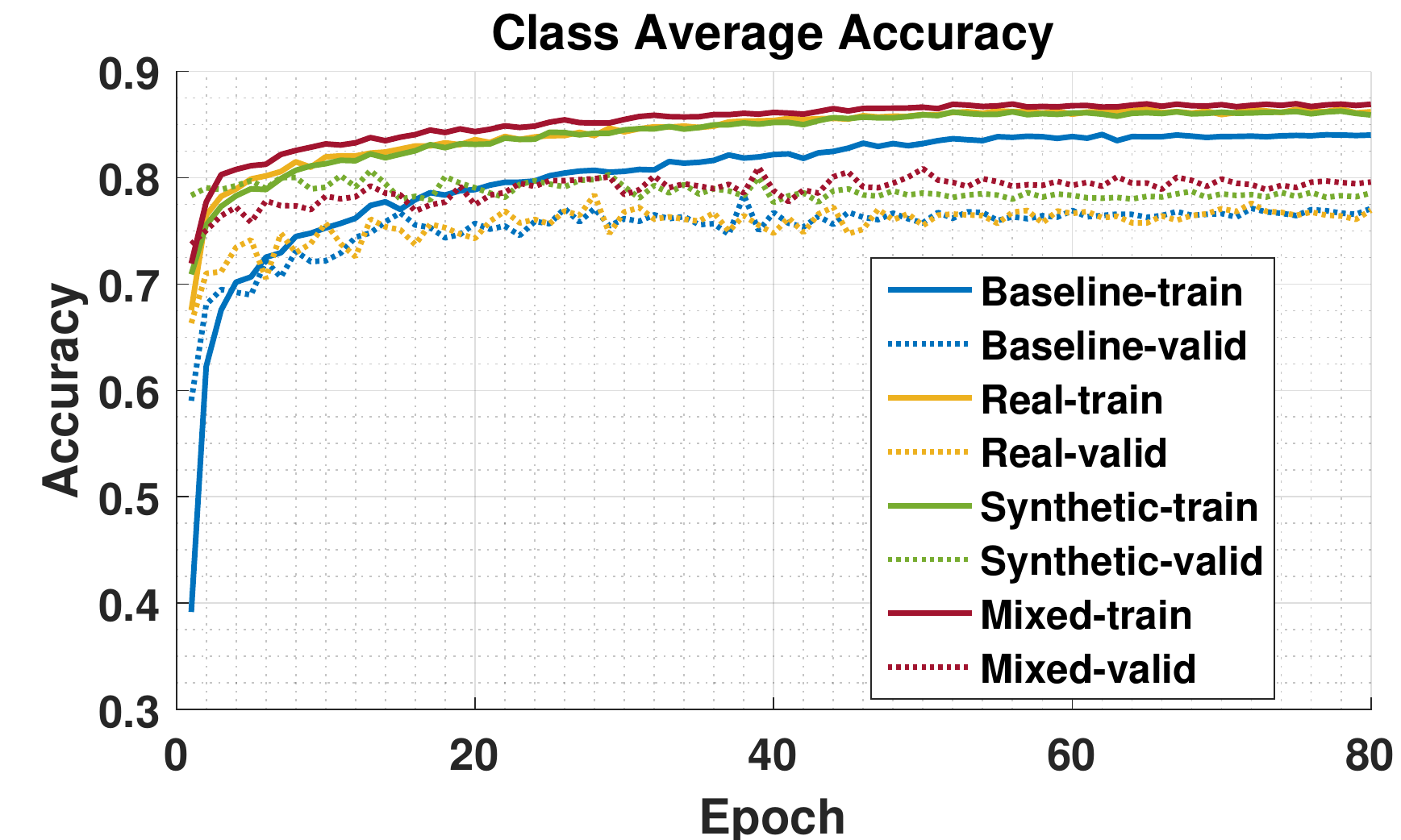}
    \caption{The influence of various pre-training approaches on the \texttt{CamVid+} dataset. The solid lines are the evaluation results on the training set, the dashed lines are the results on the \textit{validation} set.}
    \label{sup:fig:camvidp_conv}
\end{figure}

\begin{figure}
    \includegraphics[width=0.5\textwidth]{experiments/cityscapes+/objective.pdf}
    \includegraphics[width=0.5\textwidth]{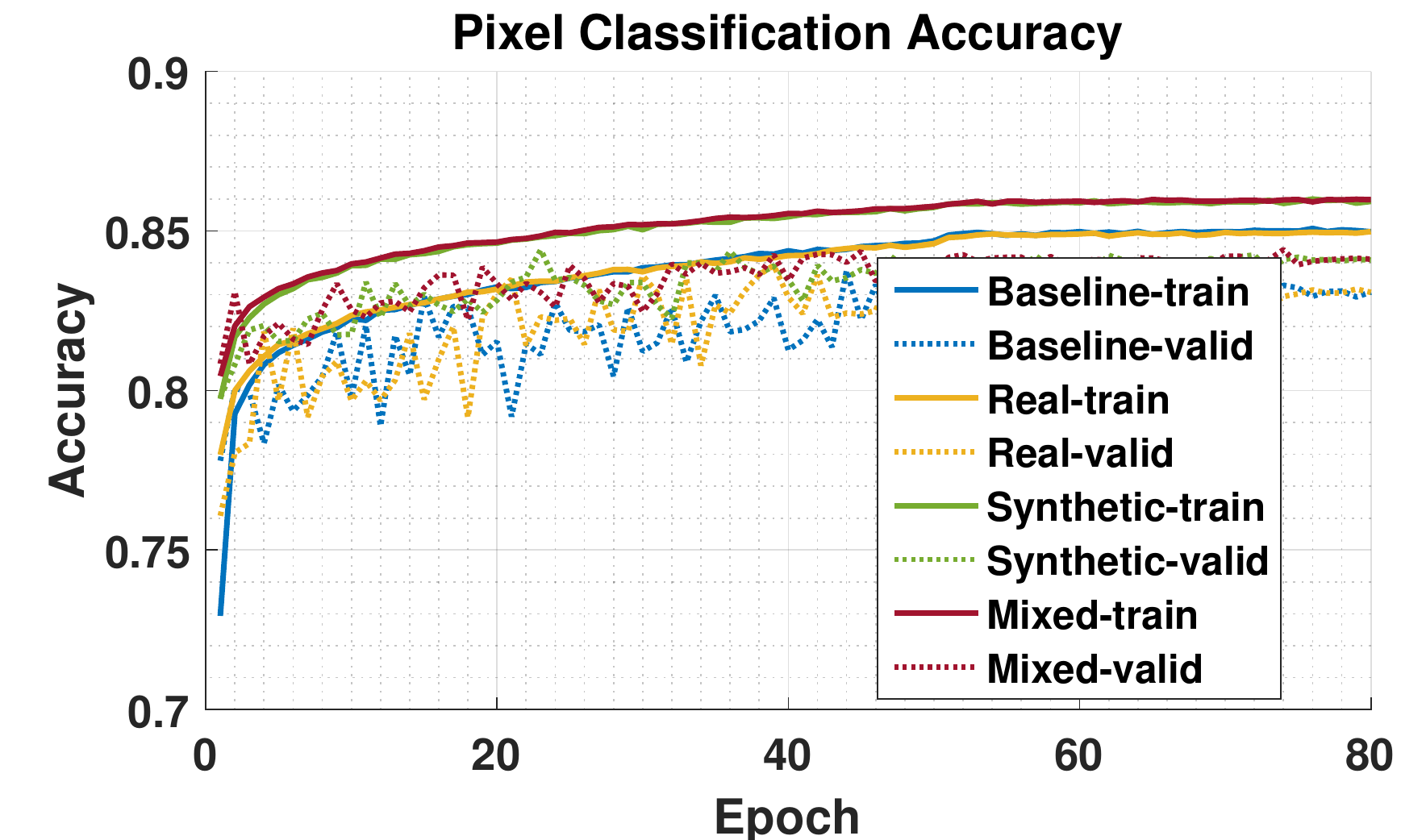}
    \includegraphics[width=0.5\textwidth]{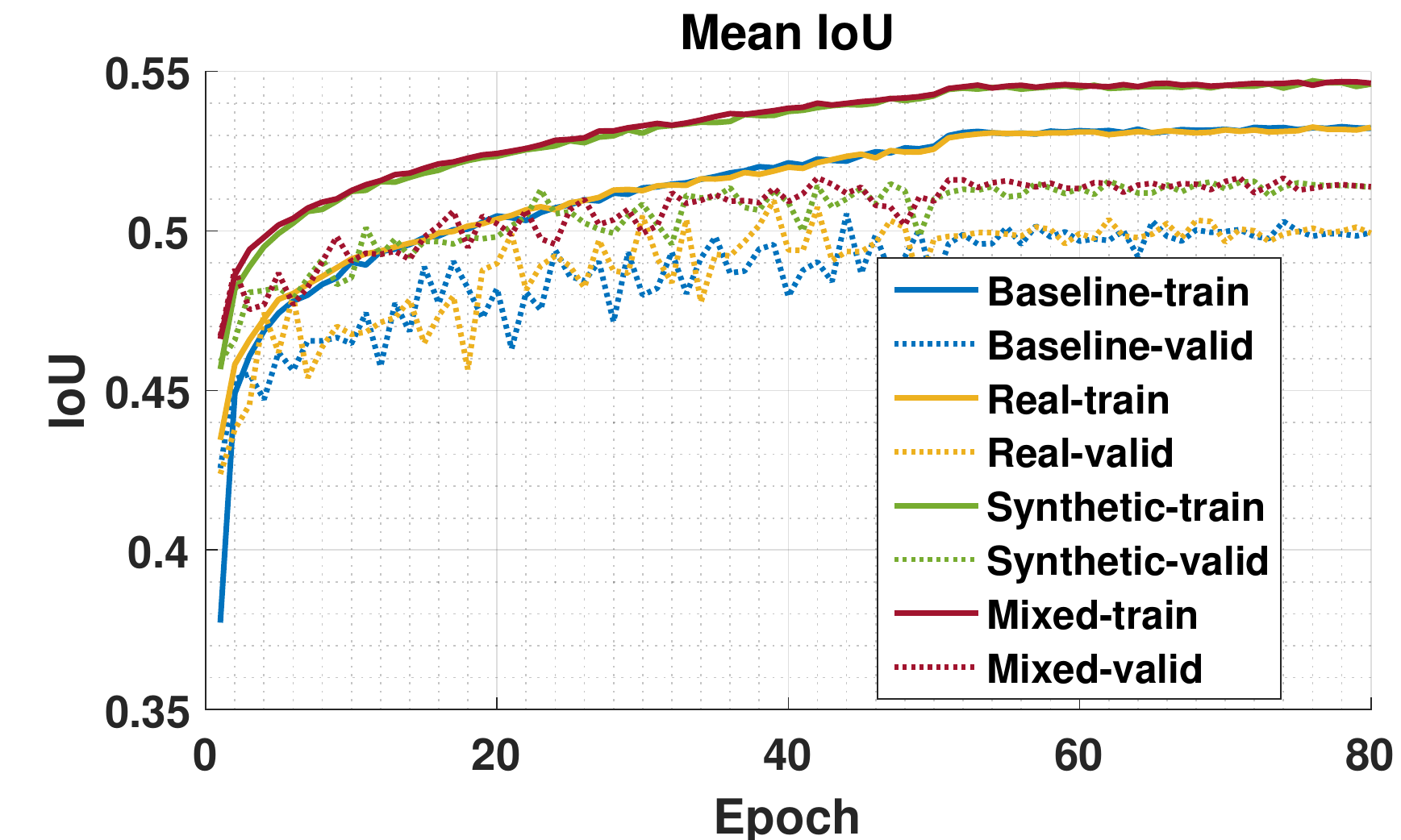}
    \includegraphics[width=0.5\textwidth]{experiments/cityscapes+/meanAcc.pdf}
    \caption{The influence of various pre-training approaches on the \texttt{Cityscapes+} dataset. The solid lines are the evaluation results on the training set, the dashed lines are the results on the \textit{validation} set.}
    \label{sup:fig:cityp_conv}
\end{figure}

Figures \ref{sup:fig:camvid_conv}-\ref{sup:fig:cityp_conv} compare the behavior of our convolutional networks in the training phase. We present the evolution of \textit{objective value}, \textit{pixel classification accuracy}, \textit{class average accuracy}, and \textit{mean intersection-over-union} for \texttt{CamVid}, \texttt{Cityscapes}, \texttt{CamVid+}, and \texttt{Cityscapes+} datasets. Pre-training on synthetic data consistently improves the initialization and the final solution, and in most cases also outperforms pre-training on real-world data.


\begin{figure}
\centering
    \includegraphics[width=0.45\textwidth]{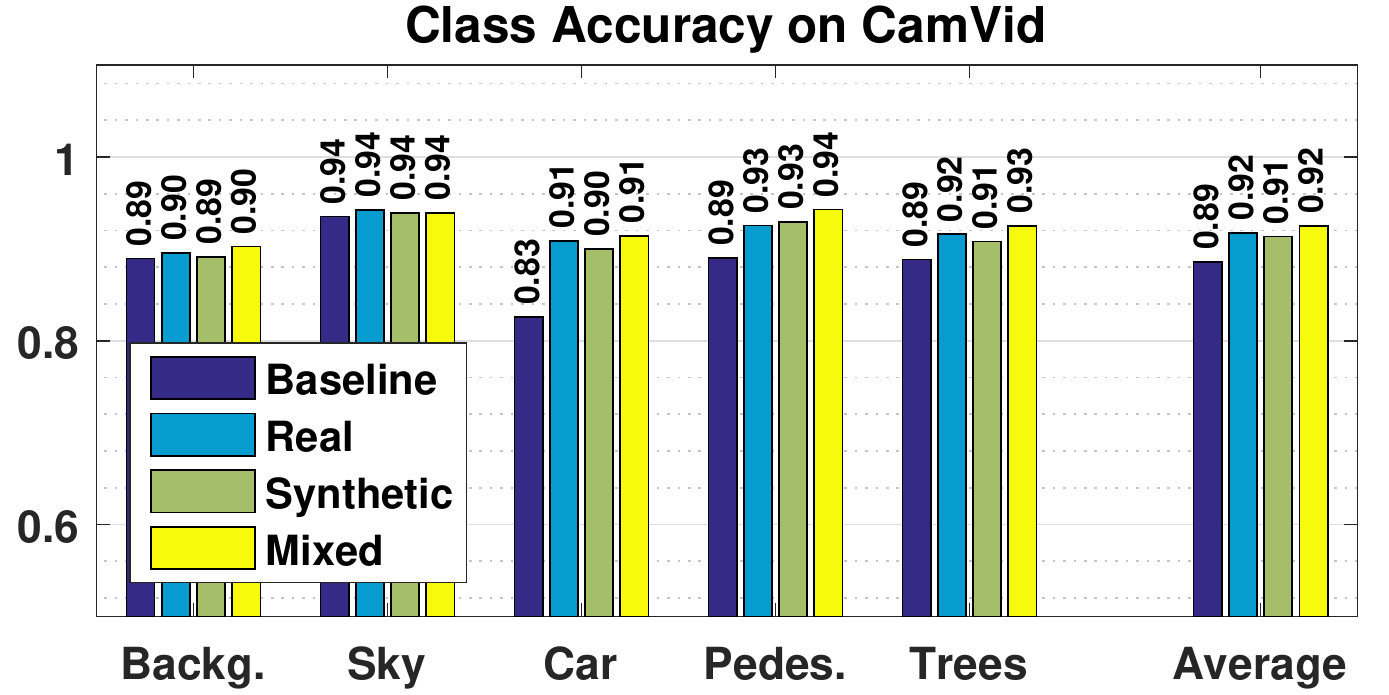}
    \includegraphics[width=0.45\textwidth]{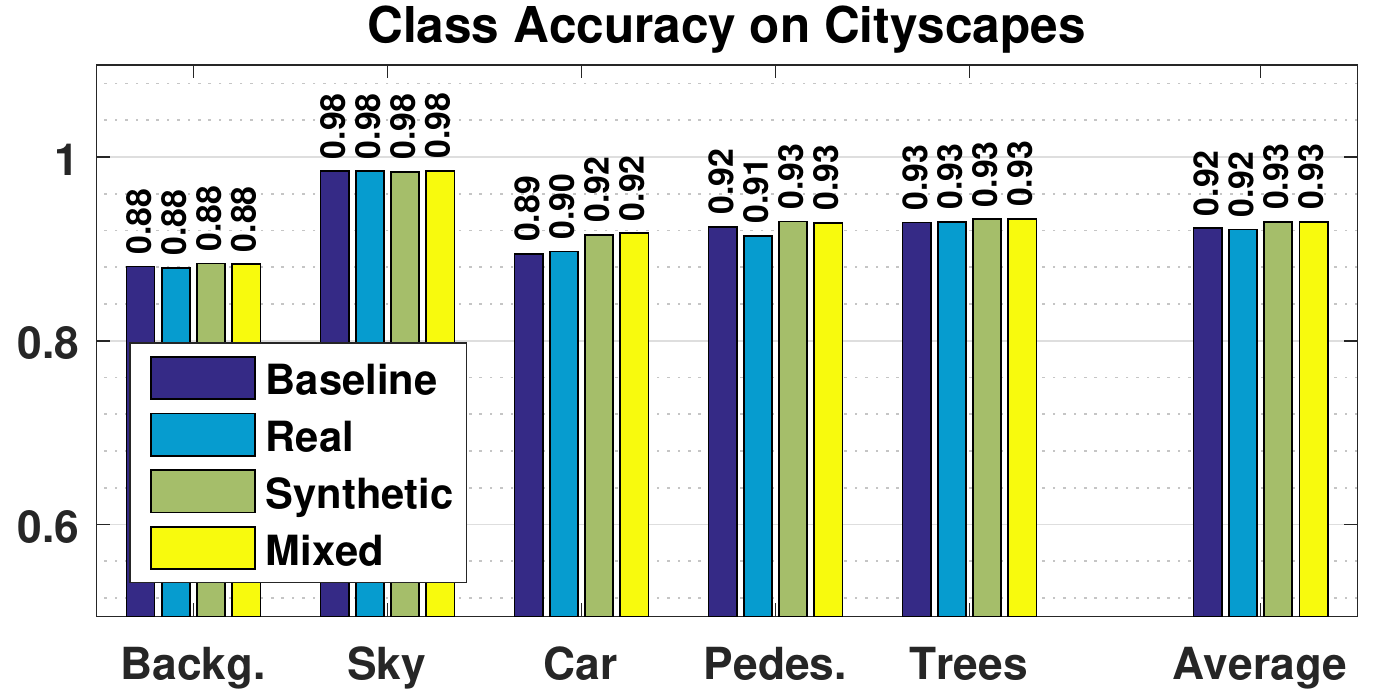}
    \caption{The per-class accuracy on \texttt{CamVid} (left) and \texttt{Cityscapes} (right).}
    \label{sup:fig:cam_ci_class}
\end{figure}

\begin{figure}
\centering
    \includegraphics[width=0.9\textwidth,trim=0.8cm 0.3cm 0cm 0cm, clip=true]{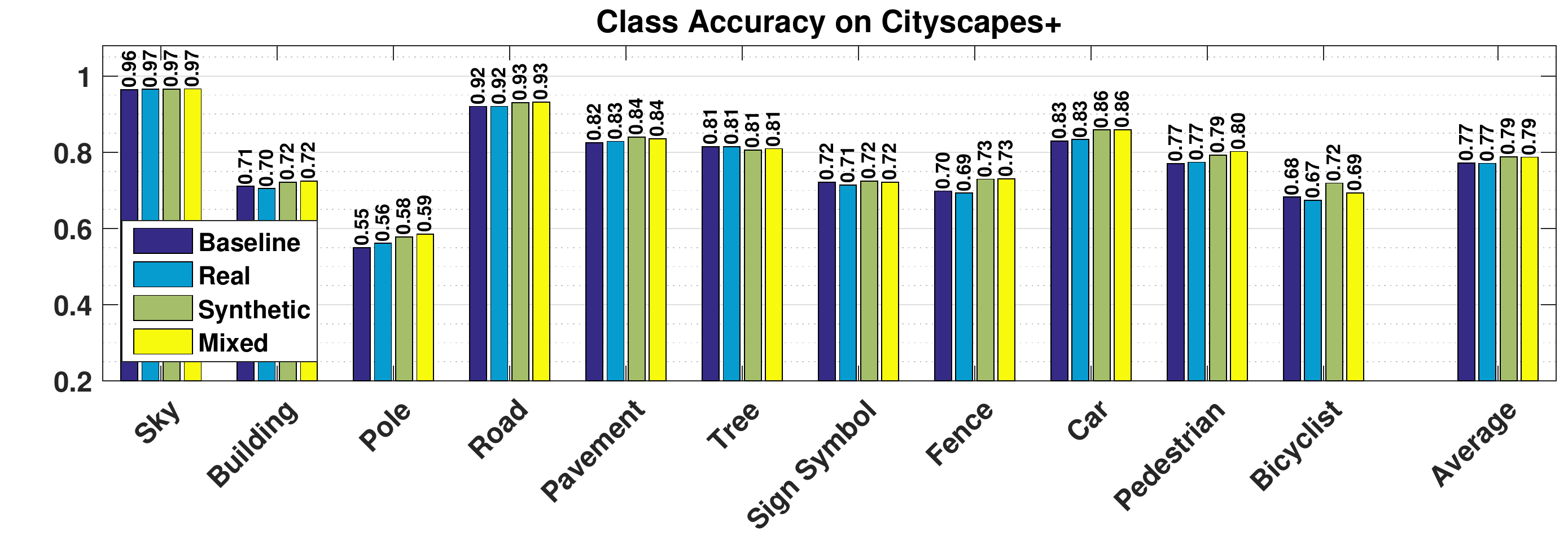}
    \caption{The per-class accuracy on \texttt{Cityscapes+}.}
    \label{sup:fig:cityp_class}
\end{figure}

Figure~\ref{sup:fig:cam_ci_class} compares the per-class accuracy of each training strategy on the test set of \texttt{CamVid} and the validation set of \texttt{Cityscapes}. Using synthetic data yields a consistent improvement over the baseline. On \texttt{CamVid}, pre-training on real data leads to a better model than pre-training on synthetic data, but the mixed approach has the best accuracy. On \texttt{Cityscapes}, however, pre-training on synthetic data has a higher average accuracy than pre-training on real-world data.
Figure~\ref{sup:fig:cityp_class} shows the per-class accuracy on the \texttt{Cityscapes+} dataset. Similar to the previous experiments using synthetic data results in more improvement than using real-world data.
Combining synthetic and real data gives the highest performance boost in these experiments.


\subsection*{Cross-dataset Evaluation}
\begin{figure}
    \includegraphics[width=\textwidth,trim=0cm 0.4cm 0cm 0cm, clip=true]{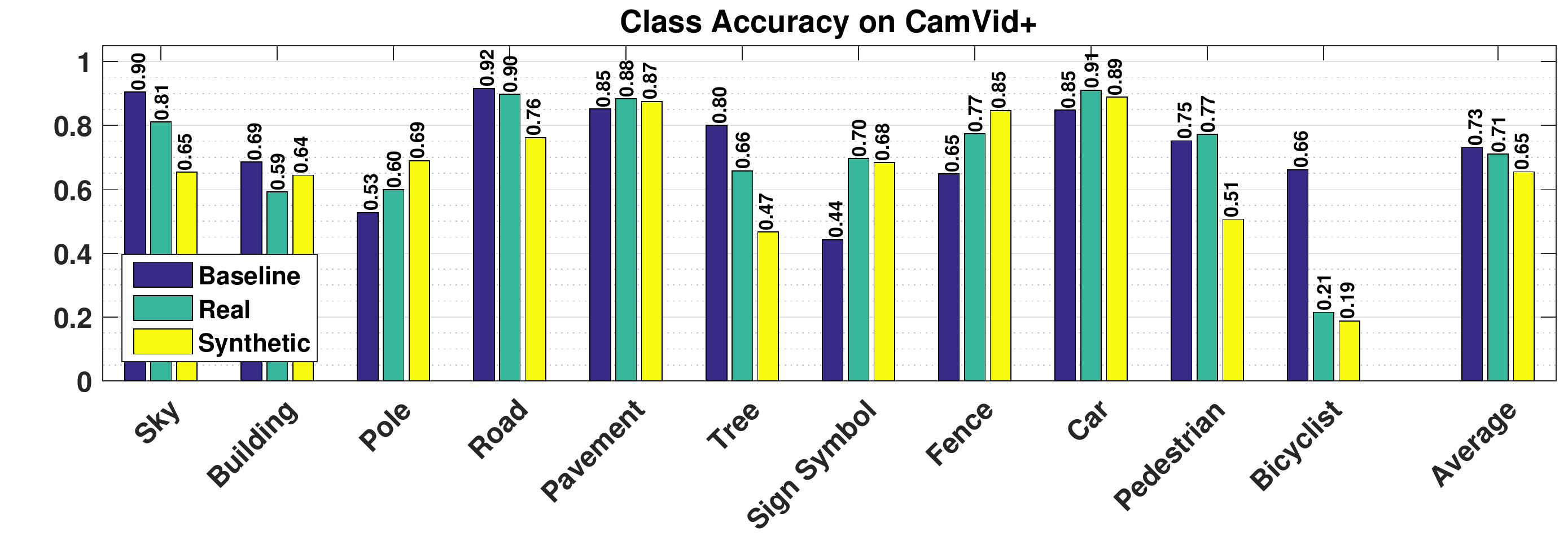}
    \caption{Cross-dataset evaluation. The per-class accuracy on the test set of the \texttt{CamVid+} dataset.}
    \label{sup:fig:camvidp_cross_class}
\end{figure}

In the cross-dataset setting, we train one network on each dataset and evaluate the accuracy of each network on the other datasets. The purpose of this experiment is to measure and compare the generalization power of the networks that are trained on synthetic or real data only. Figure \ref{sup:fig:camvidp_cross_class} shows the per-class accuracy for evaluation on the \texttt{Camvid+} dataset. The \texttt{Baseline} network is directly trained on the target dataset, while the \texttt{Real} network is trained on the alternative real dataset, and the \texttt{Synthetic} network is trained on synthetic data only. Without domain adaptation, both of the \texttt{Real} and \texttt{Synthetic} networks have a lower accuracy than the \texttt{Baseline}. The network that is trained on real data has a better accuracy than the network that is trained on synthetic data only. Even though the \texttt{Synthetic} network is only trained on synthetic data, it outperforms the real network on `Building', `Pole', and `Fence'. 
While the \texttt{Synthetic} network does not exceed the accuracy of the \texttt{Real} network on average, the small gap indicates that the network with synthetic data is relying on relevant features and is not merely overfitting to the game specific textures.

\subsection{Depth Estimation from RGB}
\begin{figure}
\centering
    \includegraphics[width=0.49\textwidth]{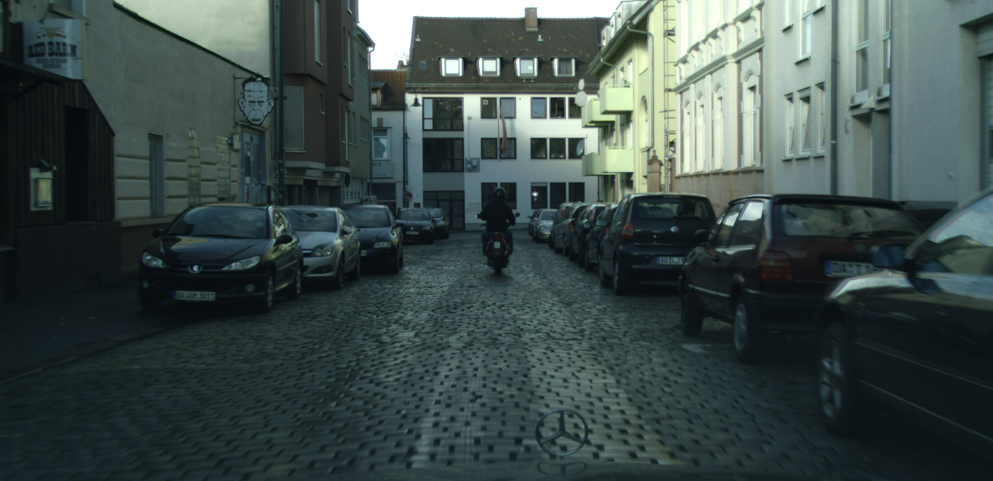}
    \includegraphics[width=0.49\textwidth]{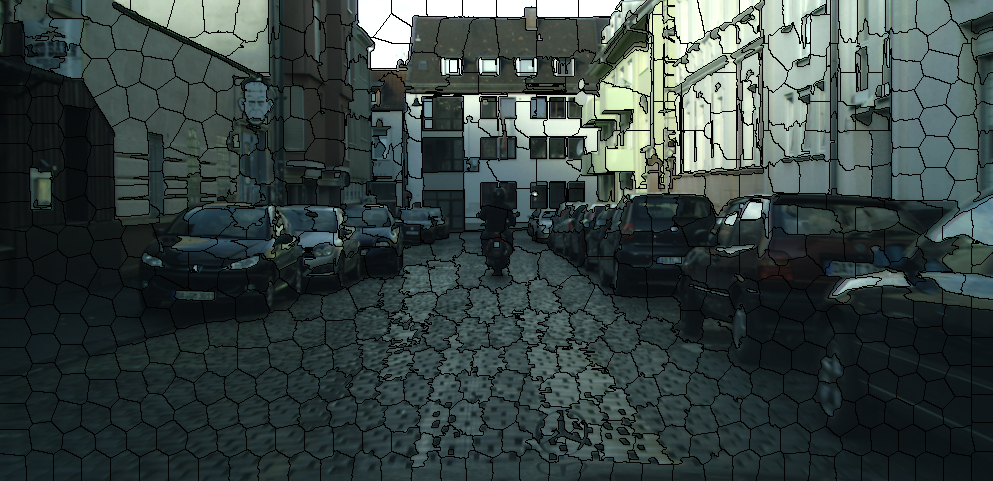}
    \\
    \includegraphics[width=0.49\textwidth]{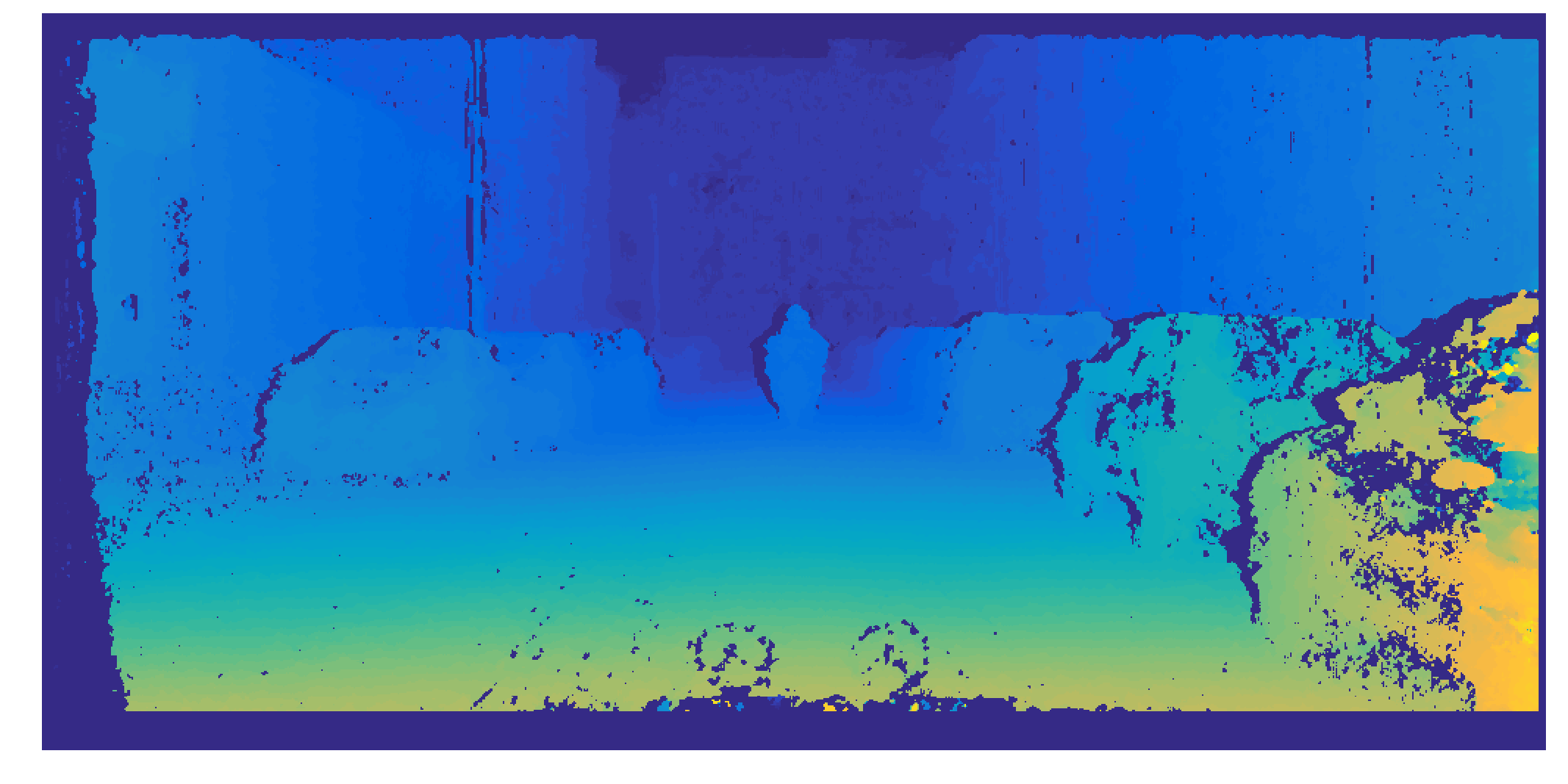}
    \includegraphics[width=0.49\textwidth]{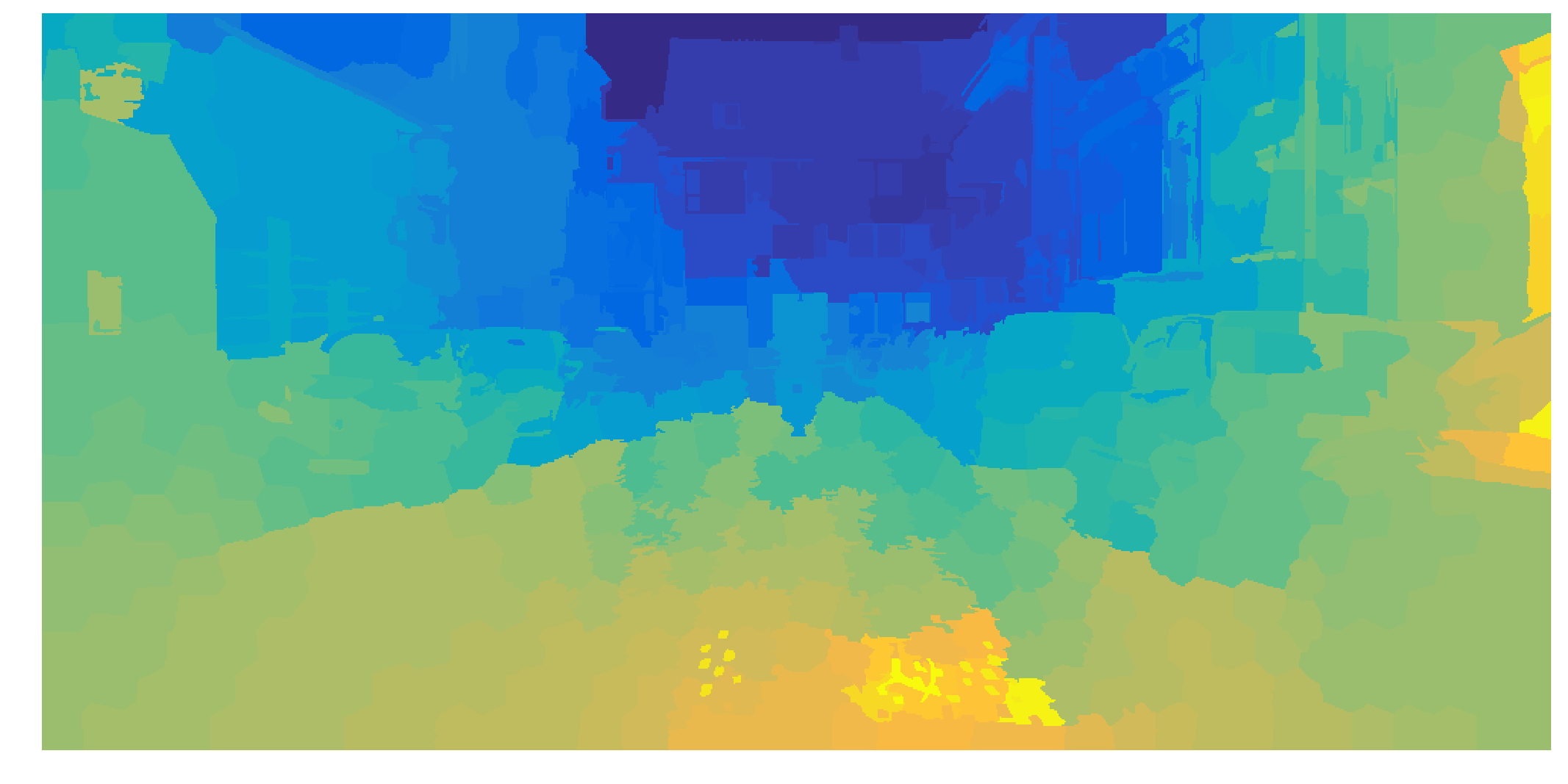}

    \caption{(top left) A sample image from the \texttt{Cityscapes}~\cite{Cordts2016Cityscapes} dataset, (top right) decomposition of the RGB image to SLIC superpixels~\cite{achanta2012slic}, (bottom left) the groundtruth disparity map, (bottom right) the globalized depth output of the method presented by Zoran~\etal~\cite{Zoran_2015_ICCV}.}
    \label{sup:fig:depth}
\end{figure}

Zoran~\etal~\cite{Zoran_2015_ICCV} present a depth estimation method that only relies on the ordinal relationships between a set of image patch pairs.
The image is first decomposed into SLIC~\cite{achanta2012slic} superpixels. A deep convolutional network classifies the ordinal relationship between
the adjacent superpixels by generating a local relationship label $\{<,\,=,\,>\}$ with the corresponding probabilities. A quadratic program is then constructed
to generate a total ordering (ranking) over the superpixels which will represent the depth.
Note that this method does not rely on the depth measurement unit. Hence, it is not directly comparable to the prior work that directly regress to the depth value.
We use this method because the depth information that is collected in the video game is not directly comparable to the real-world depth metrics.
The ordinal relationships, however, can be consistently inferred from the extracted depth information.
In the main paper, we demonstrated how using the synthetic RGB images can improve the patch classifier of Zoran~\etal~\cite{Zoran_2015_ICCV}.
Figure~\ref{sup:fig:depth} shows the groundtruth depth and the predicted depth image of a sample input from the \texttt{Cityscapes} dataset.

\end{document}